# Theoretical Analyses of Cross-Validation Error and Voting in Instance-Based Learning

Peter Turney
Knowledge Systems Laboratory
Institute for Information Technology
National Research Council Canada
Ottawa, Ontario, Canada
K1A 0R6

613-993-8564
peter@ai.iit.nrc.ca

**Running Head:** Error and Voting

# Theoretical Analyses of Cross-Validation Error and Voting in Instance-Based Learning

## Abstract

This paper begins with a general theory of error in cross-validation testing of algorithms for supervised learning from examples. It is assumed that the examples are described by attribute-value pairs, where the values are symbolic. Cross-validation requires a set of training examples and a set of testing examples. The value of the attribute that is to be predicted is known to the learner in the training set, but unknown in the testing set. The theory demonstrates that cross-validation error has two components: error on the training set (inaccuracy) and sensitivity to noise (instability).

This general theory is then applied to voting in instance-based learning. Given an example in the testing set, a typical instance-based learning algorithm predicts the designated attribute by voting among the $k$ nearest neighbors (the $k$ most similar examples) to the testing example in the training set. Voting is intended to increase the stability (resistance to noise) of instance-based learning, but a theoretical analysis shows that there are circumstances in which voting can be destabilizing. The theory suggests ways to minimize cross-validation error, by insuring that voting is stable and does not adversely affect accuracy.

## 1 Introduction

This paper is concerned with cross-validation testing of algorithms for supervised learning from examples. It is assumed that the examples are described by attribute-value pairs, where the attributes can have a finite number of symbolic values. The learning task is to predict the value of one of the attributes, given the value of the remaining attributes. It is assumed that each example is described by the same set of attributes.

Without loss of generality, we may assume that the attributes are restricted to boolean values. Let us suppose that each example has $r + 1$ attributes. We may think of an example as a boolean vector in the space $\{0,1\}^{r+1}$. The learning task is to construct a function from $\{0,1\}^r$ to $\{0,1\}$, where $\{0,1\}^r$ is the space of the predictor attributes and $\{0,1\}$ is the space of the attribute that is to be predicted.

Cross-validation testing requires a set of training examples and a set of testing examples. The learning algorithm (the *student*) is scored by a *teacher*, according to the student's performance on the testing set. The teacher knows the values of all of the attributes of all of the examples in the training set and the testing set. The student knows all of the values in the training set, but the student does not know the values of the attribute that is to be predicted in the testing set. The student uses the training set to develop a model of the data. The student then uses the model to make predictions for the testing set. The teacher compares the student's predictions with the actual values in the testing set. A mismatch between the student's prediction and the actual value is counted as an error. The student's goal is to minimize the number of errors that it makes on the testing set.

In a recent paper (Turney, 1993), a general theory of error in cross-validation testing of algorithms for predicting real-valued attributes is presented. [1] Section 2 of this paper extends the theory of Turney (1993) to algorithms for predicting boolean-valued attributes. This section shows that cross-validation error has two components: error on the training set and sensitivity to noise. Error on the training set is commonly used as a measure of *accuracy* (Fraser, 1976). Turney (1990) introduces a formal definition of *stability* as a measure of the sensitivity of algorithms to noise. Section 2 proves that cross-validation error is bounded by the sum of the training set error and the instability. The optimal cross-validation error is established, and it is proven that the strategy of minimizing error on the training set (maximizing accuracy) is sub-optimal.

Section 3 examines the cross-validation error of a simple form of instance-based learning (Aha *et al.*, 1991; Kibler *et al.*, 1989). Instance-based learning is not a single learning algorithm; it is a paradigm for a class of learning algorithms. It is related to the nearest neighbor pattern classification paradigm (Dasarathy, 1991). In instance-based learning, the student's model of the data consists of simply storing the training set. Given an example from the testing set, the student makes a prediction by looking for similar examples in the training set. Section 3 proves that this simple form of instance-based learning produces sub-optimal cross-validation error.

Looking for the single most similar example is sub-optimal because it is overly sensitive to noise in the data. Section 4 deals with instance-based learning algorithms that look for the $k$ most similar examples, where $k \geq 1$. In general, the $k$ most similar examples will not all agree on the value of the attribute that is to be predicted. This section examines algorithms that resolve this conflict by *voting* (Dasarathy, 1991). For example, suppose $k = 3$ and the value of the attribute that is to be predicted is 1 for two of the three examples and 0 for the remaining example. If the algorithm uses majority voting, then the prediction is that the value is 1.

Section 4 presents a detailed theoretical analysis of voting in instance-based learning. One motivation for voting is the desire to make instance-based learning more stable (more resistant to noise in the data). This section examines the stability of voting in the best case, the worst case, and the average case. It is shown that, in the worst case, voting can be *destabilizing*. That is, in certain circumstances, voting can actually increase the sensitivity to noise.

Section 5 discusses the practical application of this theory. It is shown how it is possible to estimate the expected cross-validation error from the training set. This section presents estimators that can indicate whether voting will increase or decrease stability for a particular set of data. This gives a method for choosing the value of $k$ that will minimize the cross-validation error.

Section 6 compares the work presented here with related work. The most closely related work is Turney (1993), which first introduced many of the concepts used here. There are interesting differences, which arise because Turney (1993) involves real-valued attributes and classes, while this paper involves boolean-valued attributes and classes. This theory is closely related to Akaike Information Criterion statistics (Sakamoto *et al.*, 1986), as is discussed elsewhere (Turney, 1993). There is also an interesting connection with some prior work in nearest neighbor pattern classification (Cover & Hart, 1967).

Finally, Section 7 considers future work. One weakness of this general theory of cross-validation error is that it does not model interpolation and extrapolation. Another area for future research is applying the theory to problems other than voting in instance-based learning.

# 2 Cross-Validation Error

This section presents a general theory of error in cross-validation testing of algorithms for predicting symbolic attributes. In order to make the exposition simpler, the discussion is restricted to boolean-valued attributes. It is not difficult to extend the results presented here to $n$-valued attributes, where $n$ is any integer larger than one.

## 2.1 Exclusive-Or

This paper makes extensive use of the boolean exclusive-or operator. Suppose $x$ and $y$ are boolean variables. That is, $x$ and $y$ range over the set $\{0,1\}$. We may write $x \oplus y$ for "$x$ exclusive-or $y$". The expression $x \oplus y$ has the value 1 if and only if exactly one of $x$ and $y$ has the value 1. Here are some of the properties of exclusive-or:

$$x \oplus 0 = x \tag{1}$$

$$x \oplus 1 = \neg x \tag{2}$$

$$x \oplus y = y \oplus x \tag{3}$$

$$x \oplus x = 0 \tag{4}$$

$$x \oplus \neg x = 1 \tag{5}$$

$$x \oplus (y \oplus z) = (x \oplus y) \oplus z \tag{6}$$

These properties can easily be verified with a truth-table.

The following results use exclusive-or with boolean vectors. Suppose $\vec{x}$ and $\vec{y}$ are boolean vectors in the space $\{0,1\}^n$. The expression $\vec{x} \oplus \vec{y}$ represents the vector that results from applying exclusive-or to corresponding elements of $\vec{x}$ and $\vec{y}$:

$$\vec{z} = \vec{x} \oplus \vec{y} \tag{7}$$

$$\vec{x} = \begin{bmatrix} x_1 & \dots & x_n \end{bmatrix} \tag{8}$$

$$\vec{y} = \begin{bmatrix} y_1 & \dots & y_n \end{bmatrix} \tag{9}$$

$$\vec{z} = \begin{bmatrix} x_1 \oplus y_1 & \dots & x_n \oplus y_n \end{bmatrix} \tag{10}$$

## 2.2 Accuracy and Stability

Suppose we have a black box with $r$ inputs and one output, where the inputs and the output can be represented by boolean values. Let the boolean vector $\vec{v}$ represent the inputs to the black box:

$$\vec{v} = \begin{bmatrix} x_1 & \dots & x_r \end{bmatrix} \tag{11}$$

Let $y$ represent the output of the black box. Suppose that the black box has a deterministic component $f$ and a random component $z$. The deterministic component $f$ is a function that maps from $\{0,1\}^r$ to $\{0,1\}$. The random component $z$ is a random boolean variable. The probability that $z$ is 1 is $p$. The probability that $z$ is 0 is $1 - p$. The black box is represented by the equation:

$$y = f(\vec{v}) \oplus z \tag{12}$$

The variable $p$ is a probability in the range $[0,1]$.

If $p$ is close to 0, then $z$ is usually 0, so the deterministic component dominates and the output $y$ is usually $f(\vec{v})$ (see equation (1) above). If $p$ is close to 1, then $z$ is usually 1, so the output is usually $\neg f(\vec{v})$ (see (2)). We may think of $p$ as the probability that $f(\vec{v})$ will be randomly negated. If $p$ is 0.5, then the random component $z$ completely hides the deterministic component $f$. When $p$ is between 0.5 and 1.0, $f(\vec{v})$ will be negated more often than not. When we know that $p$ is greater than 0.5, we can negate the output of the black box:

$$y = \neg(f(\vec{v}) \oplus z) \tag{13}$$

This counteracts the expected negation of $f(\vec{v})$ by the random component $z$. In (12) the output $y$ is $f(\vec{v})$ with probability $1 - p$. In (13) the output $y$ is $f(\vec{v})$ with probability $p$.

Suppose we perform a series of $n$ experiments with the black box. We may represent the inputs in the $n$ experiments with a matrix $X$:

$$X = \begin{bmatrix} x_{1,1} & \dots & x_{1,r} \\ \dots & x_{i,j} & \dots \\ x_{n,1} & \dots & x_{n,r} \end{bmatrix} \tag{14}$$

Let the $i$-th row of the matrix $X$ be represented by the vector $\vec{v}_i$, where:

$$\vec{v}_i = \begin{bmatrix} x_{i,1} & \dots & x_{i,r} \end{bmatrix} \qquad i = 1, \dots, n \tag{15}$$

The vector $\vec{v}_i$ contains the values of the $r$ inputs for the $i$-th experiment. Let the $j$-th column of the matrix $X$ be represented by the vector $\vec{x}_j$, where:

$$\vec{x}_j = \begin{bmatrix} x_{1,j} \\ \dots \\ x_{n,j} \end{bmatrix} \qquad j = 1, \dots, r \tag{16}$$

The vector $\vec{x}_j$ contains the values of the $j$-th input for the $n$ experiments. Let the $n$ outputs be represented by the vector $\vec{y}$, where:

$$\vec{y} = \begin{bmatrix} y_1 \\ \dots \\ y_n \end{bmatrix} \tag{17}$$

The scalar $y_i$ is the output of the black box for the $i$-th experiment.

The function $f$ can be extended to a vector function $\vec{f}(X)$, where:

$$\vec{f}(X) = \begin{bmatrix} f(\vec{v}_1) \\ \dots \\ f(\vec{v}_n) \end{bmatrix} \tag{18}$$

Our model for the $n$ experiments is:

$$\vec{y} = \vec{f}(X) \oplus \vec{z} \tag{19}$$

The vector $\vec{z}$ is a sequence of $n$ independent random boolean variables, each having the value 1 with probability $p$:

$$\vec{z} = \begin{bmatrix} z_1 \\ \dots \\ z_n \end{bmatrix} \tag{20}$$

Let us suppose that we are trying to develop a model of $f$. We may write $m(\vec{v} | X, \vec{y})$ to represent the model's prediction for $f(\vec{v})$, given that the model is based on the data $X$ and $\vec{y}$. We can extend $m(\vec{v} | X, \vec{y})$ to a vector function:

$$\vec{m}(X | X, \vec{y}) = \begin{bmatrix} m(\vec{v}_1 | X, \vec{y}) \\ \dots \\ m(\vec{v}_n | X, \vec{y}) \end{bmatrix} \tag{21}$$

Thus $\vec{m}(X | X, \vec{y})$ represents the model's prediction for $\vec{f}(X)$, given the data $X$ and $\vec{y}$.

Suppose we repeat the whole sequence of $n$ experiments, holding the inputs $X$ constant:

$$\vec{y}_1 = \grave{f}(X) \oplus \vec{z}_1 \quad (22)$$

$$\vec{y}_2 = \grave{f}(X) \oplus \vec{z}_2 \quad (23)$$

The outputs of the first set of *n* experiments are represented by $\vec{y}_1$ and the outputs of the second set of *n* experiments are represented by $\vec{y}_2$:

$$\vec{y}_1 = \begin{bmatrix} y_{1,1} \\ \dots \\ y_{1,n} \end{bmatrix} \qquad \vec{y}_2 = \begin{bmatrix} y_{2,1} \\ \dots \\ y_{2,n} \end{bmatrix} \quad (24)$$

Although the inputs *X* are the same, the outputs may have changed, due to the random component. That is, assuming *p* is neither 0 nor 1, it is possible that $\vec{z}_1 \neq \vec{z}_2$:

$$\vec{z}_1 = \begin{bmatrix} z_{1,1} \\ \dots \\ z_{1,n} \end{bmatrix} \qquad \vec{z}_2 = \begin{bmatrix} z_{2,1} \\ \dots \\ z_{2,n} \end{bmatrix} \quad (25)$$

Let the data $(X, \vec{y}_1)$ be the training set and let the data $(X, \vec{y}_2)$ be the testing set in cross-validation testing of the model *m*. The cross-validation error vector $\vec{e}_c$ is:

$$\vec{e}_c = \vec{m}(X | X, \vec{y}_1) \oplus \vec{y}_2 \quad (26)$$

$$\vec{e}_c = \begin{bmatrix} e_{c,1} \\ \dots \\ e_{c,n} \end{bmatrix} = \begin{bmatrix} m(\vec{v}_1 | X, \vec{y}_1) \oplus y_{2,1} \\ \dots \\ m(\vec{v}_n | X, \vec{y}_1) \oplus y_{2,n} \end{bmatrix} \quad (27)$$

$e_{c,i}$ is 1 if and only if the prediction of the model $m(\vec{v}_i | X, \vec{y}_1)$ differs from the output $y_{2,i}$ in the testing data. We may write $\|\vec{e}_c\|$ for the length of the cross-validation error vector:

$$\|\vec{e}_c\| = \sum_{i=1}^{n} e_{c,i} \quad (28)$$

In other words, $\|\vec{e}_c\|$ is the number of errors that the model $\vec{m}(X | X, \vec{y}_1)$ makes on the testing set.

The model $m(\vec{v} | X, \vec{y}_1)$ has been tuned to perform well on the training set $(X, \vec{y}_1)$. Therefore the number of errors made on the training set may be deceptively low. To get a good indication of the quality of the model, we must test it on an independent set of data. Some authors call this error measure "cross-validation error", while other authors call it "train-and-test error" (Weiss and Kulikowski, 1991). In this paper, the former terminology is used. [2]

It is assumed that our goal is to minimize the expected number of errors $E(\|\vec{e}_c\|)$ that the model makes on the testing set. $E(\ldots)$ is the expectation operator from probability theory (Fraser, 1976). If $t(x)$ is a function of a random variable $x$, where $x \in S$ ($S$ is the set of possible values of $x$) and the probability of observing a particular value of $x$ is $p(x)$, then $E(t(x))$ is:

$$E(t(x)) = \sum_{x \in S} t(x)p(x) \tag{29}$$

The expected cross-validation error $E(\|\vec{e}_c\|)$ depends on the random boolean vectors $\vec{z}_1$ and $\vec{z}_2$.

The next theorem gives some justification for the assumption that $X$ is the same in the training set $(X, \vec{y}_1)$ and the testing set $(X, \vec{y}_2)$.

**Theorem 1:** Let $D$ be an arbitrary probability distribution on $\{0,1\}^r$. Let the $n$ rows $\vec{v}_1, \ldots, \vec{v}_n$ of $X$ be independently randomly selected according to the distribution $D$. Let the vector $\vec{v}$ be randomly selected according to the distribution $D$. Let $p^*$ be the probability that $\vec{v}$ is not equal to any of $\vec{v}_1, \ldots, \vec{v}_n$. Then:

$$E(p^*) \le (1 - 2^{-r})^n \tag{30}$$

*Proof:* Let $D(\vec{v})$ be the probability of randomly selecting $\vec{v}$ with the distribution $D$. The probability that $\vec{v}$ is not equal to any of $\vec{v}_1, \ldots, \vec{v}_n$ is thus:

$$p^* = (1 - D(\vec{v}))^n \tag{31}$$

The expected value of $p^*$ is:

$$E(p^*) = \sum_{\vec{v} \in \{0,1\}^r} p^* D(\vec{v}) \tag{32}$$

$$= \sum_{\vec{v} \in \{0,1\}^r} (1 - D(\vec{v}))^n D(\vec{v}) \tag{33}$$

The expected value is at its maximum when $D$ is the uniform distribution on $\{0,1\}^r$. For the uniform distribution, $D(\vec{v}) = 2^{-r}$. Therefore:

$$E(p^*) \le \sum_{\vec{v} \in \{0,1\}^r} (1 - 2^{-r})^n 2^{-r} \tag{34}$$

$$= 2^r (1 - 2^{-r})^n 2^{-r} \tag{35}$$

$$= (1 - 2^{-r})^n \tag{36}$$

$\square$.

The implication of Theorem 1 is that, as $n$ increases, the probability that the inputs in the training set and the testing set are the same approaches 1. Thus the assumption that $X$ is the same in the training set and the testing set is reasonable for large values of $n$.

Although it is assumed that the inputs are boolean, only Theorem 1 uses this assumption; none of the following results depend on the assumption that the inputs are boolean. The inputs could just as well be real numbers, so that $f$ maps from $\Re^r$ to $\{0,1\}$. However, it is necessary to assume that the output is boolean. If the output is real-valued, then we may turn to the analysis in Turney (1993).

Cover and Hart (1967) prove the following Lemma (their symbols have been changed to be consistent with the notation used here):

**Lemma 1:** Let $\vec{v}$ and $\vec{v}_1, \ldots, \vec{v}_n$ be independent identically distributed random variables taking values in a separable metric space $S$, with metric $d$. Let $\vec{v}_i$ be the nearest neighbor, according to the metric $d$, to $\vec{v}$ in the set $\vec{v}_1, \ldots, \vec{v}_n$. Then, as $n$ approaches infinity, $d(\vec{v}, \vec{v}_i)$ approaches 0, with probability 1.

*Proof:* See Cover and Hart (1967) $\square$.

Cover and Hart (1967) assume that $n$ is large and the probability distribution for $P(f(\vec{v}) \oplus z = 1)$ is a continuous function of $\vec{v}$. It follows from Lemma 1 that, as $n$ approaches infinity, $P(f(\vec{v}) \oplus z = 1)$ approaches $P(f(\vec{v}_i) \oplus z = 1)$. This is similar to Theorem 1 here, except that $S$ is continuous. Note that, if the probability distribution for $P(f(\vec{v}) \oplus z = 1)$ is a continuous function of $\vec{v}$, then $z$ must be a function of $\vec{v}$, $z(\vec{v})$, in order to compensate for the fact that $f(\vec{v})$ is discontinuous. Thus the assumption that the distribution is continuous is relatively strong.

For the rest of this paper, let us assume that $f$ maps from $\{0,1\}^r$ to $\{0,1\}$. We may assume that $X$ is the same in the training set and the testing set. This assumption may be justified with Theorem 1. Alternatively, we could assume that $f$ maps from $\Re^r$ to $\{0,1\}$ and use Lemma 1 to justify the assumption that $X$ is the same in the training set and the testing set. However, the analysis is simpler when $z$ is not a function of $\vec{v}$.

The error on the training set $\vec{e}_t$ is:

$$\vec{e}_t = \vec{m}(X|X,\vec{y}_1) \oplus \vec{y}_1 \tag{37}$$

$$\vec{e}_t = \begin{bmatrix} e_{t,1} \\ \dots \\ e_{t,n} \end{bmatrix} = \begin{bmatrix} m(\vec{v}_1|X,\vec{y}_1) \oplus y_{1,1} \\ \dots \\ m(\vec{v}_n|X,\vec{y}_1) \oplus y_{1,n} \end{bmatrix} \tag{38}$$

$e_{t,i}$ is 1 if and only if the prediction of the model $m(\vec{v}_i|X,\vec{y}_1)$ differs from the output $y_{1,i}$ in the training data.

There is another form of error that is of interest. We may call this error *instability*:

$$\vec{e}_s = \vec{m}(X|X,\vec{y}_1) \oplus \vec{m}(X|X,\vec{y}_2) \tag{39}$$

$$\vec{e}_s = \begin{bmatrix} e_{s,1} \\ \dots \\ e_{s,n} \end{bmatrix} = \begin{bmatrix} m(\vec{v}_1|X,\vec{y}_1) \oplus m(\vec{v}_1|X,\vec{y}_2) \\ \dots \\ m(\vec{v}_n|X,\vec{y}_1) \oplus m(\vec{v}_n|X,\vec{y}_2) \end{bmatrix} \tag{40}$$

Instability is a measure of the sensitivity of the model to noise in the data. If the model resists noise, then $\vec{m}(X|X,\vec{y}_1)$ will be relatively similar to $\vec{m}(X|X,\vec{y}_2)$, so $E(\|\vec{e}_s\|)$ will be relatively small. If the model is sensitive to noise, then $\vec{m}(X|X,\vec{y}_1)$ will be relatively dissimilar from $\vec{m}(X|X,\vec{y}_2)$, so $E(\|\vec{e}_s\|)$ will be relatively large.

The following theorem shows the relationship between cross-validation error, error on the training set, and instability.

**Theorem 2:** The expected cross-validation error is less than or equal to the sum of the expected error on the training set and the expected instability:

$$E(\|\vec{e}_c\|) \le E(\|\vec{e}_t\|) + E(\|\vec{e}_s\|) \tag{41}$$

*Proof:* Let us introduce a new term $\vec{e}_\omega$:

$$\vec{e}_\omega = \vec{m}(X|X,\vec{y}_2) \oplus \vec{y}_1 \tag{42}$$

Due to the symmetry of the training set (22) and the testing set (23), we have (from (26) and (42)):

$$E(\|\vec{e}_c\|) = E(\|\vec{e}_\omega\|) \tag{43}$$

Recall the definition of $\vec{e}_t$ (37):

$$\vec{e}_t = \vec{m}(X|X,\vec{y}_1) \oplus \vec{y}_1 \tag{44}$$

It follows from this definition that $\vec{y}_1$ differs from $\vec{m}(X|X,\vec{y}_1)$ in $\|\vec{e}_t\|$ locations. Recall the definition of $\vec{e}_s$ (39):

$$\vec{e}_s = \vec{m}(X|X,\vec{y}_1) \oplus \vec{m}(X|X,\vec{y}_2) \tag{45}$$

It follows from this definition that $\vec{m}(X|X,\vec{y}_1)$ differs from $\vec{m}(X|X,\vec{y}_2)$ in $\|\vec{e}_s\|$ locations. Therefore $\vec{y}_1$ differs from $\vec{m}(X|X,\vec{y}_2)$ in at most $\|\vec{e}_t\| + \|\vec{e}_s\|$ locations. Thus:

$$\|\vec{e}_\omega\| \le \|\vec{e}_t\| + \|\vec{e}_s\| \tag{46}$$

Finally:

$$E(\|\vec{e}_c\|) = E(\|\vec{e}_\omega\|) \le E(\|\vec{e}_t\| + \|\vec{e}_s\|) = E(\|\vec{e}_t\|) + E(\|\vec{e}_s\|) \tag{47}$$

□ .

The next theorem is a variation on Theorem 2.

**Theorem 3:** If $\vec{e}_t$ and $\vec{e}_s$ are statistically independent, then:

$$E(\|\vec{e}_c\|) = E(\|\vec{e}_t\|) + E(\|\vec{e}_s\|) - \frac{2}{n}E(\|\vec{e}_t\|)E(\|\vec{e}_s\|) \tag{48}$$

*Proof:* Assume that $\vec{e}_t$ and $\vec{e}_s$ are statistically independent. Let us introduce a new term

$\vec{e}_\omega$:

$$\vec{e}_\omega = \vec{m}(X|X,\vec{y}_2) \oplus \vec{y}_1 \tag{49}$$

We see that:

$$E(\|\vec{e}_c\|) = E(\|\vec{e}_\omega\|) \tag{50}$$

Recall the definition of $\vec{e}_t$ (37):

$$\vec{e}_t = \vec{m}(X|X,\vec{y}_1) \oplus \vec{y}_1 \tag{51}$$

Recall the definition of $\vec{e}_s$ (39):

$$\vec{e}_s = \vec{m}(X|X,\vec{y}_1) \oplus \vec{m}(X|X,\vec{y}_2) \tag{52}$$

We have:

$$\vec{e}_t \oplus \vec{e}_s = (\vec{m}(X|X,\vec{y}_1) \oplus \vec{y}_1) \oplus (\vec{m}(X|X,\vec{y}_1) \oplus \vec{m}(X|X,\vec{y}_2)) \tag{53}$$

$$= (\vec{m}(X|X,\vec{y}_1) \oplus \vec{m}(X|X,\vec{y}_1)) \oplus (\vec{y}_1 \oplus \vec{m}(X|X,\vec{y}_2)) \tag{54}$$

$$= \vec{y}_1 \oplus \vec{m}(X|X,\vec{y}_2) \tag{55}$$

$$= \vec{e}_\omega \tag{56}$$

Let us introduce the following terms:

$$p_t = E(\|\vec{e}_t\|)/n \tag{57}$$

$$p_s = E(\|\vec{e}_s\|)/n \tag{58}$$

$$p_\omega = E(\|\vec{e}_\omega\|)/n \tag{59}$$

Since $\vec{e}_t$ and $\vec{e}_s$ are statistically independent and $\vec{e}_\omega = \vec{e}_t \oplus \vec{e}_s$, we have:

$$p_\omega = p_t(1-p_s) + p_s(1-p_t) = p_t + p_s - 2p_t p_s \tag{60}$$

Therefore:

$$E(\|\vec{e}_c\|) = E(\|\vec{e}_\omega\|) \tag{61}$$

$$= np_\omega \tag{62}$$

$$= n(p_t + p_s - 2p_t p_s) \tag{63}$$

$$= E(\|\vec{e}_t\|) + E(\|\vec{e}_s\|) - \frac{2}{n}E(\|\vec{e}_t\|)E(\|\vec{e}_s\|) \tag{64}$$

□ .

The assumptions of Theorem 3 are stronger than the assumptions of Theorem 2, but Theorem 3 gives a closer relationship between $E(\|\vec{e}_c\|)$, $E(\|\vec{e}_t\|)$, and $E(\|\vec{e}_s\|)$.

The next theorem shows the optimal expected cross-validation error.

**Theorem 4:** Suppose that *f* and *p* are known to the modeler. To minimize the expected cross-validation error, the modeler should set the model as follows:

$$\vec{m}(X|\, X,\vec{y}_i) = \begin{bmatrix} \vec{f}(X) & \text{if } p \le 0.5 \\ \neg\vec{f}(X) & \text{if } p > 0.5 \end{bmatrix} \tag{65}$$

With this model, we have:

$$E(\|\vec{e}_s\|) = 0 \tag{66}$$

$$E(\|\vec{e}_c\|) = E(\|\vec{e}_t\|) = \begin{bmatrix} np & \text{if } p \le 0.5 \\ n(1-p) & \text{if } p > 0.5 \end{bmatrix} \tag{67}$$

*Proof:* Suppose that $p \le 0.5$. By (4), (39), and (65):

$$E(\|\vec{e}_s\|) = E(\|\vec{f}(X) \oplus \vec{f}(X)\|) = 0 \tag{68}$$

Thus the model is perfectly stable. This is natural, since the model is not based on the data; it is based on the *a priori* knowledge of *f*. Consider the cross-validation error:

$$\vec{e}_c = \vec{f}(X) \oplus \vec{y}_2 \tag{69}$$

$$= \vec{f}(X) \oplus (\vec{f}(X) \oplus \vec{z}_2) \tag{70}$$

$$= (\vec{f}(X) \oplus \vec{f}(X)) \oplus \vec{z}_2 \tag{71}$$

$$= \vec{z}_2 \tag{72}$$

Thus:

$$E(\|\vec{e}_c\|) = E(\|\vec{z}_2\|) = E\left(\sum_{i=1}^{n} z_{2,i}\right) = \sum_{i=1}^{n} E(z_{2,i}) = \sum_{i=1}^{n} p = np \tag{73}$$

Consider the error on the training set:

$$\vec{e}_t = \vec{f}(X) \oplus \vec{y}_1 = \vec{z}_1 \tag{74}$$

Therefore:

$$E(\|\vec{e}_t\|) = E(\|\vec{z}_1\|) = E(\|\vec{z}_2\|) = E(\|\vec{e}_c\|) = np \tag{75}$$

Now, suppose that $p > 0.5$. For instability, we have:

$$E(\|\vec{e}_s\|) = E(\|\neg\grave{f}(X) \oplus \neg\grave{f}(X)\|) = 0 \tag{76}$$

Consider the cross-validation error:

$$\vec{e}_c = \neg\grave{f}(X) \oplus \vec{y}_2 \tag{77}$$

$$= \neg\grave{f}(X) \oplus (\grave{f}(X) \oplus \vec{z}_2) \tag{78}$$

$$= (\neg\grave{f}(X) \oplus \grave{f}(X)) \oplus \vec{z}_2 \tag{79}$$

$$= \neg\vec{z}_2 \tag{80}$$

Thus:

$$E(\|\vec{e}_c\|) = E(\|\neg\vec{z}_2\|) = E\left(\sum_{i=1}^{n} \neg z_{2,i}\right) = \sum_{i=1}^{n} E(\neg z_{2,i}) = n(1-p) \tag{81}$$

Consider the error on the training set:

$$\vec{e}_t = \neg\grave{f}(X) \oplus \vec{y}_1 = \neg\vec{z}_1 \tag{82}$$

Therefore:

$$E(\|\vec{e}_t\|) = E(\|\neg\vec{z}_1\|) = E(\|\neg\vec{z}_2\|) = E(\|\vec{e}_c\|) = n(1-p) \tag{83}$$

It is clear that no model can have a lower $E(\|\vec{e}_c\|)$ than this model, since $\vec{z}_2$ cannot be predicted $\square$.

The next theorem considers models that minimize the error on the training set.

**Theorem 5:** Let our model be as follows:

$$\vec{m}(X|X,\vec{y}_i) = \vec{y}_i \tag{84}$$

Then we have:

$$E(\|\vec{e}_t\|) = 0 \tag{85}$$

$$E(\|\vec{e}_c\|) = E(\|\vec{e}_s\|) = 2np - 2np^2 \tag{86}$$

*Proof:* Consider the expected error on the training set:

$$E(\|\vec{e}_t\|) = E(\|\vec{m}(X|X,\vec{y}_1) \oplus \vec{y}_1\|) = E(\|\vec{y}_1 \oplus \vec{y}_1\|) = 0 \tag{87}$$

Consider the cross-validation error:

$$\vec{e}_c = \vec{m}(X|X,\vec{y}_1) \oplus \vec{y}_2 \quad (88)$$

$$= \vec{y}_1 \oplus \vec{y}_2 \quad (89)$$

$$= (\vec{f}(X) \oplus \vec{z}_1) \oplus (\vec{f}(X) \oplus \vec{z}_2) \quad (90)$$

$$= (\vec{f}(X) \oplus \vec{f}(X)) \oplus (\vec{z}_1 \oplus \vec{z}_2) \quad (91)$$

$$= \vec{z}_1 \oplus \vec{z}_2 \quad (92)$$

Therefore:

$$E(\|\vec{e}_c\|) = E(\|\vec{z}_1 \oplus \vec{z}_2\|) \quad (93)$$

$$= E\left(\sum_{i=1}^{n} z_{1,i} \oplus z_{2,i}\right) \quad (94)$$

$$= \sum_{i=1}^{n} E(z_{1,i} \oplus z_{2,i}) \quad (95)$$

$$= 2np - 2np^2 \quad (96)$$

Consider the expected instability:

$$E(\|\vec{e}_s\|) = E(\|\vec{m}(X|X,\vec{y}_1) \oplus \vec{m}(X|X,\vec{y}_2)\|) = E(\|\vec{y}_1 \oplus \vec{y}_2\|) = E(\|\vec{e}_c\|) \quad (97)$$

□ .

These theorems are variations on theorems that first appeared in Turney (1993). Theorems 2, 3, 4, and 5 in this paper correspond to Theorems 1, 12, 2, and 3 in Turney (1993). The difference is that the theorems in Turney (1993) cover real-valued variables, while the theorems here cover boolean-valued variables. A comparison will show that there are interesting contrasts between the continuous case and the discrete case.

We shall call the model of Theorem 4 $m_\alpha$:

$$\vec{m}_\alpha(X|X,\vec{y}_i) = \begin{bmatrix} \vec{f}(X) & \text{if } p \le 0.5 \\ \neg\vec{f}(X) & \text{if } p > 0.5 \end{bmatrix} \quad (98)$$

We shall call the model of Theorem 5 $m_\beta$:

$$\vec{m}_\beta(X|X,\vec{y}_i) = \vec{y}_i \quad (99)$$

Let us compare $E(\|\vec{e}_c\|)$ for $m_\alpha$ and $m_\beta$. Let $p_\alpha$ be $E(\|\vec{e}_c\|)/n$ for $m_\alpha$:

$$p_\alpha = \begin{bmatrix} p & \text{if } p \le 0.5 \\ (1-p) & \text{if } p > 0.5 \end{bmatrix} \tag{100}$$

Let $p_\beta$ be $E(\|\grave{e}_c\|)/n$ for $m_\beta$:

$$p_\beta = 2p - 2p^2 \tag{101}$$

Figure 1 is a plot of $E(\|\grave{e}_c\|)/n$ as a function of *p*, for $m_\alpha$ and $m_\beta$. $m_\alpha$ has lower expected cross-validation error than $m_\beta$, except at the points 0, 0.5, and 1, where $m_\alpha$ has the same expected cross-validation error as $m_\beta$. When *p* equals 0 or 1, there is no noise in the black box, so $m_\alpha$ and $m_\beta$ have the same expected cross-validation error. When *p* equals 0.5, the noise completely hides the deterministic component *f*, so again $m_\alpha$ and $m_\beta$ have the same expected cross-validation error. In general, when *p* is neither 0, 0.5, nor 1, models that minimize error on the training set (such as $m_\beta$) will give sub-optimal cross-validation error.

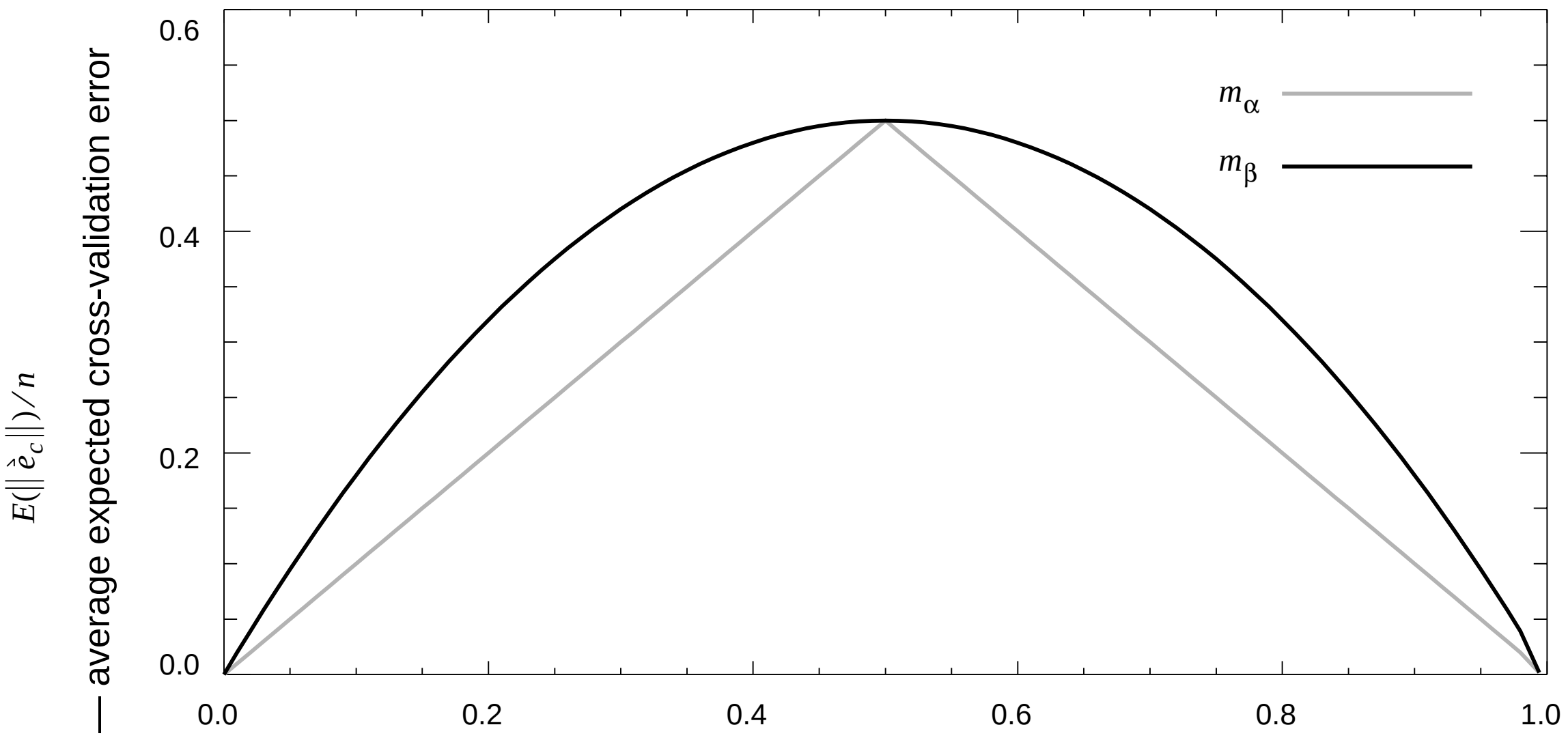

Figure 1. Plot of $E(\|\grave{e}_c\|)/n$ as a function of *p*.

The next theorem shows the relation between $p_\alpha$ and $p_\beta$.

**Theorem 6:** For $0 \le p \le 1$:

$$p_\beta = 2p_\alpha - 2p_\alpha^2 \tag{102}$$

*Proof:* When $0 \le p \le 0.5$, $p_\alpha = p$. Therefore we have:

$$p_\beta = 2p_\alpha - 2p_\alpha^2 \tag{103}$$

When $0.5 < p \le 1$, $p_\alpha = 1 - p$. Therefore we have:

$$2p_\alpha - 2p_\alpha^2 = 2(1-p) - 2(1-p)^2 \tag{104}$$

$$= 2 - 2p - 2 + 4p - 2p^2 \tag{105}$$

$$= p_\beta \tag{106}$$

$\square$.

Theorem 6 shows that, for small values of *p*, $p_\beta$ is approximately twice the optimal $p_\alpha$.

At this point, it may be worthwhile to summarize the assumptions that are made in this theory:

1. It is assumed that the inputs (the features; the attributes; the matrix *X*) are the same in the training set $(X, \vec{y}_1)$ and the testing set $(X, \vec{y}_2)$. This assumption is the weakest element in this theory. It is discussed in detail earlier in this section and also in Section 7. It is also discussed in more depth in Turney (1993). Note that the outputs $\vec{y}_1$ and $\vec{y}_2$ are *not* the same in the training and testing sets.

2. It is assumed that the inputs are boolean-valued. This assumption is only used in Theorem 1. The purpose of Theorem 1 is to increase the credibility of the above assumption 1. Theorem 1 can easily be adapted to the case of multi-valued symbolic attributes. Lemma 1 covers the case of real-valued attributes. Thus there is nothing essential about the assumption that the inputs are boolean-valued.

3. It is assumed that the noise vector $\vec{z}$ is a sequence of *n* independent random boolean variables, each having the value 1 with probability *p*. That is, $\vec{z}$ is a sequence of samples from a Bernoulli(*p*) distribution (Fraser, 1976). This is a very weak assumption, which is likely to be (approximately) satisfied in most real-world data (given assumption 6 below).

4. It is assumed that there is noise in the class attribute (assumption 3 above), but noise in the input attributes is not addressed (due to assumption 1 above).

5. It is assumed that the data consist of a deterministic component $f$ and a random component $z$. This assumption is expressed in the model $y = f(\vec{v}) \oplus z$. One implication of this assumption is that the target concept $f$ does not drift with time or shift with context.

6. It is assumed that the output $y$ (the class attribute) is boolean-valued. The theory can easily be extended to handle multi-valued symbolic class attributes. Turney (1993) discusses real-valued outputs (and real-valued noise vectors).

With the exception of the first assumption, these assumptions are relatively weak. [3]

This section has presented some general results that apply to any algorithm for predicting boolean-valued attributes. The rest of this paper focuses on a particular class of algorithms: instance-based learning algorithms. This section has presented a general theory and the remainder of the paper will demonstrate that the theory can be fruitfully applied to a real, concrete machine learning algorithm.

# 3 Single Nearest Neighbor

Instance-based learning may be used either for predicting boolean-valued attributes (Aha *et al.*, 1991) or for predicting real-valued attributes (Kibler *et al.*, 1989). Instance-based learning is a paradigm for a class of learning algorithms; it is not a single algorithm. It is related to the nearest neighbor pattern recognition paradigm (Dasarathy, 1991).

With instance-based learning, the model $m(\vec{v} | X, \vec{y})$ is constructed by simply storing the data $(X, \vec{y})$. These stored data are the *instances*. In order to make a prediction for the input $\vec{v}$, we examine the row vectors $\vec{v}_1, \ldots, \vec{v}_n$ of the matrix $X$. In the simplest version of instance-based learning, we look for the row vector $\vec{v}_i$ that is most similar to the input $\vec{v}$. The prediction for the output is $m(\vec{v} | X, \vec{y}) = y_i$, the element of $\vec{y}$ that corresponds to the row vector $\vec{v}_i$.

There are many ways that one might choose to measure the similarity between two vectors. It is assumed only that we are using a reasonable measure of similarity. Let us say that a similarity measure $\mathrm{sim}(\vec{u}_1, \vec{u}_2)$ is *reasonable* if:

$$\vec{u}_1 \neq \vec{u}_2 \rightarrow \text{sim}(\vec{u}_1, \vec{u}_2) < \text{sim}(\vec{u}_1, \vec{u}_1) \tag{107}$$

That is, the similarity between distinct vectors is always less than the similarity between identical vectors.

The $k$ row vectors in $X$ that are most similar to the input $\vec{v}$ are called the *k nearest neighbors* of $\vec{v}$ (Dasarathy, 1991). We may use $m_k$ to represent instance-based learning with $k$ nearest neighbors. This section focuses on $k = 1$.

The next theorem shows that instance-based learning $m_1$ using the single nearest neighbor gives sub-optimal cross-validation error.

**Theorem 7:** Let the model $m_1$ use instance-based learning with the single nearest neighbor. That is, if the row vector $\vec{v}_i$ is the most similar to the input $\vec{v}$ of all the row vectors $\vec{v}_1,\ldots,\vec{v}_n$ of the matrix $X$, then $m_1(\vec{v}|X,\vec{y}) = y_i$. If $m_1$ uses a reasonable measure of similarity and no two rows in $X$ are identical, then $\vec{m}_1(X|X,\vec{y}) = \vec{y}$.

*Proof:* Since $m_1$ is based on a reasonable measure of similarity and no two rows in $X$ are identical, no vector is more similar to $\vec{v}_i$ than $\vec{v}_i$ itself, so $m_1(\vec{v}_i|X,\vec{y}) = y_i$. Thus it follows from (21) that $\vec{m}(X|X,\vec{y}) = \vec{y}$ □.

Theorem 7 shows that $m_1$ satisfies the assumptions of Theorem 5. Therefore $m_1$ has the following properties:

$$E(\|\vec{e}_t\|) = 0 \tag{108}$$

$$E(\|\vec{e}_c\|) = E(\|\vec{e}_s\|) = 2np - 2np^2 \tag{109}$$

In other words, $m_1$ is equivalent to $m_\beta$. We know from Section 2 that $m_1$ gives sub-optimal cross-validation error.

If there are two (or more) identical rows in $X$, then (109) still holds, but (108) is not necessarily true. $m_1$ will, in general, no longer be equivalent to $m_\beta$. The proof is a small variation on Theorem 5.

The problem with $m_1$ is that it is unstable. It is natural to consider increasing the stability of instance-based learning by considering $k > 1$.

# 4 Voting

This section examines $k \geq 1$. Suppose the model $m_k$ is given the input vector $\vec{v}$. Let $\vec{v}_1, \ldots, \vec{v}_k$ be the $k$ nearest neighbors to $\vec{v}$. That is, let $\vec{v}_1, \ldots, \vec{v}_k$ be the $k$ row vectors in $X$ that are most similar to the input vector $\vec{v}$. Let us assume a reasonable measure of similarity. Let $y_1, \ldots, y_k$ be the outputs corresponding to the rows $\vec{v}_1, \ldots, \vec{v}_k$. The model $m_k$ predicts the output for the input $\vec{v}$ to be the value of the majority of $y_1, \ldots, y_k$ (Fix & Hodges, 1951):

$$m_k(\vec{v} | X, \vec{y}) = \begin{bmatrix} 0 \text{ if } \sum_{i=1}^{k} y_i < k/2 \\ 1 \text{ if } \sum_{i=1}^{k} y_i > k/2 \end{bmatrix} \tag{110}$$

Let us assume that $k$ is an odd number, $k = 3, 5, 7, \ldots$, so there will never be a tie in majority voting with $y_1, \ldots, y_k$. Of course, we require that $k \leq n$. In general, $k \ll n$.

More generally, let us consider the following model (Tomek, 1976):

$$m_k(\vec{v} | X, \vec{y}) = \begin{bmatrix} 0 \text{ if } \sum_{i=1}^{k} y_i < tk \\ 1 \text{ if } \sum_{i=1}^{k} y_i > tk \end{bmatrix} \tag{111}$$

In this model, $t$ is a threshold, such that $0 < t < 1$. With majority voting, $t = 0.5$. When $t \neq 0.5$, it may be appropriate to consider values of $k$ other than the odd numbers.

The motivation for voting is the belief that it will increase stability; that it will make the model more resistant to noise in the data. The following sections give a formal analysis of the stability of instance-based learning with voting. They examine the best case, the worst case, and the average case.

## 4.1 Best Case

The next theorem concerns the best case, when stability is maximal.

**Theorem 8:** Let $N_k(\vec{v}_i)$ be the set of the $k$ nearest neighbors $\vec{v}_1, \ldots, \vec{v}_k$ to $\vec{v}_i$. Let $\rho_i$ be

defined as follows:

$$\rho_i = \frac{1}{k}\left(\sum_{\vec{v}_j \in N_k(\vec{v}_i)} f(\vec{v}_j)\right) \qquad i = 1, \ldots, n \tag{112}$$

That is, $\rho_i$ is the frequency with which $f$ takes the value 1, in the neighborhood $N_k(\vec{v}_i)$ of $\vec{v}_i$. Assume that we are using majority voting, $t = 0.5$, and that $k = 3, 5, 7, \ldots$. Suppose that:

$$(\forall i)\ ((\rho_i = 0) \vee (\rho_i = 1)) \tag{113}$$

This is the most stable situation; other values of $\rho_i$ are less stable. $E(\|\vec{e}_s\|)$ for $m_k$ is determined by the following formula:

$$E(\|\vec{e}_s\|) = 2nP_k - 2nP_k^2 \tag{114}$$

In this formula, $P_k$ depends on $m_k$. $P_k$ has the following values for $m_1$, $m_3$, $m_5$, and $m_\infty$:

$$P_1 = p \tag{115}$$

$$P_3 = 3p^2 - 2p^3 \tag{116}$$

$$P_5 = 6p^5 - 15p^4 + 10p^3 \tag{117}$$

$$P_\infty = \begin{bmatrix} 0 & \text{if } p < 0.5 \\ 0.5 & \text{if } p = 0.5 \\ 1 & \text{if } p > 0.5 \end{bmatrix} \tag{118}$$

*Proof:* Let $\vec{v}_1, \ldots, \vec{v}_k$ be the $k$ nearest neighbors to $\vec{v}$. Let $y_{1,1}, \ldots, y_{1,k}$ be the corresponding elements in $\vec{y}_1$. Let $y_{2,1}, \ldots, y_{2,k}$ be the corresponding elements in $\vec{y}_2$. The stability of $m_k$ is determined by the probability that:

$$\sum_{i=1}^{k} y_{1,i} < k/2 \qquad \text{and} \qquad \sum_{i=1}^{k} y_{2,i} > k/2 \tag{119}$$

If we can find this probability for each row vector in $X$, then we can find the stability of $m_k$. Suppose that $p_i$ is the probability of (119) when the input to $m_k$ is $v_i$. Then $2p_i$ is the probability that:

$$m_k(\vec{v}_i | X, \vec{y}_1) \oplus m_k(\vec{v}_i | X, \vec{y}_2) = 1 \tag{120}$$

Therefore:

$$E(\|\vec{e}_s\|) = \sum_{i=1}^{n} E(m_k(\vec{v}_i | X, \vec{y}_1) \oplus m_k(\vec{v}_i | X, \vec{y}_2)) = \sum_{i=1}^{n} 2p_i \tag{121}$$

The best case — the case with minimal expected instability $E(\|\vec{e}_s\|)$ — arises when $\rho_i = 0$ or $\rho_i = 1$. Without loss of generality, we may assume that $\rho_i = 0$. In this case, we have:

$$y_{i,j} = f(\vec{v}_j) \oplus z_{i,j} = 0 \oplus z_{i,j} = z_{i,j} \qquad i = 1, 2 \qquad j = 1, \ldots, k \tag{122}$$

Therefore (119) becomes:

$$\sum_{i=1}^{k} z_{1,i} < k/2 \qquad \text{and} \qquad \sum_{i=1}^{k} z_{2,i} > k/2 \tag{123}$$

Let us introduce the term $P_k$:

$$P_k = P\left(\sum_{i=1}^{k} z_{1,i} > k/2\right) \tag{124}$$

That is, $P_k$ is the probability that the majority of $y_{1,1}, \ldots, y_{1,k}$ are 1, given that $f(\vec{v}_1), \ldots, f(\vec{v}_k)$ are all 0. The probability of (119) is:

$$P_k(1 - P_k) \tag{125}$$

If the best case holds for every row vector in $X$ — that is, if (113) is true — then we have:

$$E(\|\vec{e}_s\|) = \sum_{i=1}^{n} 2P_k(1 - P_k) = 2nP_k(1 - P_k) \tag{126}$$

Let us consider the case $k = 3$. We have:

$$P_3 = p^3 + 3p^2(1 - p) = p^3 + 3p^2 - 3p^3 = 3p^2 - 2p^3 \tag{127}$$

Recall that for $k = 1$ we have (109):

$$E(\|\vec{e}_s\|) = 2np - 2np^2 \tag{128}$$

Thus $P_1 = p$. For $k = 5$, we have:

$$P_5 = 6p^5 - 15p^4 + 10p^3 \tag{129}$$

To find the behavior of the model in the limit, $m_\infty$, we can use the Central Limit Theorem (Fraser, 1976). Consider the following sum:

$$\sum_{i=1}^{k} z_{1,i} \tag{130}$$

The mean and variance of this sum are:

$$E\left(\sum_{i=1}^{k} z_{1,i}\right) = kp \tag{131}$$

$$\mathrm{var}\left(\sum_{i=1}^{k} z_{1,i}\right) = kp(1-p) \tag{132}$$

Consider the following expression:

$$\frac{\sum_{i=1}^{k} z_{1,i} - kp}{\sqrt{kp(1-p)}} \tag{133}$$

By the Central Limit Theorem, the distribution of (133) approaches a standard normal distribution as $k$ approaches infinity. Recall the definition of $P_k$:

$$P_k = P\left(\sum_{i=1}^{k} z_{1,i} > k/2\right) \tag{134}$$

Let $z_\alpha$ be a random variable with a standard normal distribution. As $k$ approaches infinity, $P_k$ approaches:

$$P\left(z_\alpha > \frac{(k/2) - kp}{\sqrt{kp(1-p)}}\right) \tag{135}$$

We see that:

$$\lim_{k \to \infty}\left(\frac{(k/2) - kp}{\sqrt{kp(1-p)}}\right) = \begin{bmatrix} \infty & \text{if } p < 0.5 \\ 0 & \text{if } p = 0.5 \\ -\infty & \text{if } p > 0.5 \end{bmatrix} \tag{136}$$

Therefore:

$$P_\infty = \begin{bmatrix} 0 & \text{if } p < 0.5 \\ 0.5 & \text{if } p = 0.5 \\ 1 & \text{if } p > 0.5 \end{bmatrix} \tag{137}$$

$$\frac{E(\|\grave{e}_s\|)}{n} = 2P_\infty(1 - P_\infty) = \begin{bmatrix} 0 & \text{if } p \neq 0.5 \\ 0.5 & \text{if } p = 0.5 \end{bmatrix} \tag{138}$$

□.

Theorem 8 implies that, in the best case, $m_k$ grows increasingly stable as $k$ increases, unless $p = 0.5$. Figure 2 is a plot of $E(\|\grave{e}_s\|)/n$ as a function of *p*, for the models $m_1$, $m_3$, $m_5$, and $m_\infty$.

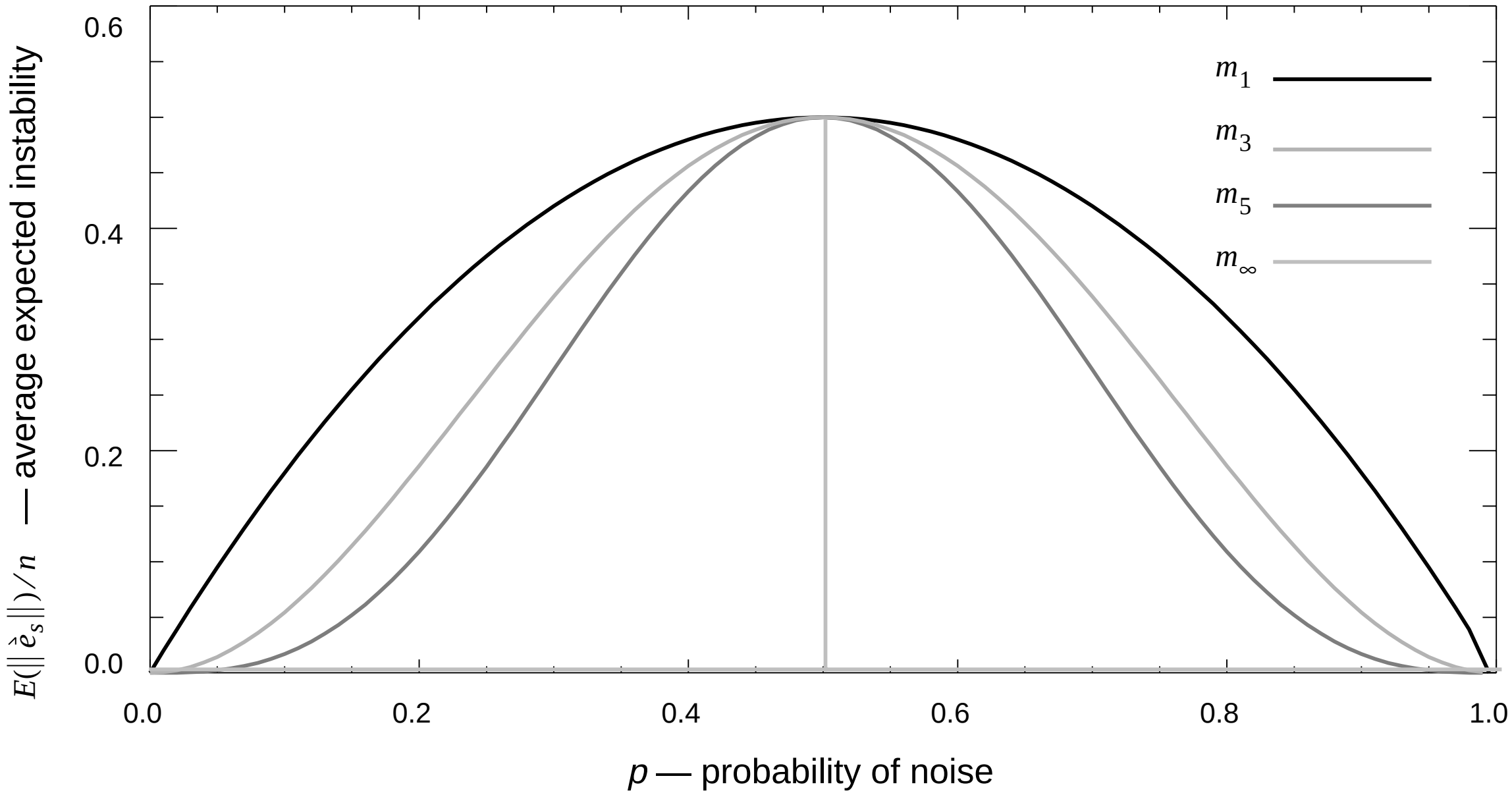

Figure 2. A plot of $E(\|\grave{e}_s\|)/n$ as a function of *p*.

When comparing Figure 1 with Figure 2, note that Figure 1 is a plot of $E(\|\grave{e}_c\|)/n$, while Figure 2 is a plot of $E(\|\grave{e}_s\|)/n$. In general, we cannot plot $E(\|\grave{e}_c\|)/n$ for $m_k$, because $E(\|\grave{e}_t\|)$ will depend on the data. Theorems 2 and 3 show that $E(\|\grave{e}_c\|)$ depends on both $E(\|\grave{e}_s\|)$ and $E(\|\grave{e}_t\|)$. The exception is $m_1$, where we know that $E(\|\grave{e}_t\|) = 0$.

(Recall that $m_1 = m_\beta$.)

Theorem 8 does not necessarily mean that we should make $k$ as large as possible, even if we assume that the best case (113) holds. It is assumed that our goal is to minimize the expected cross-validation error. Theorems 2 and 3 show that the expected cross-validation error has two components, the error on the training set and the instability. In the best case, increasing $k$ will increase stability, but increasing $k$ is also likely to increase the error on the training set. We must find the value of $k$ that best balances the conflicting demands of accuracy (low error on the training set) and stability (resistance to noise). This value will, of course, depend on $f$ and $p$.

## 4.2 Worst Case

The next theorem concerns the worst case, when stability is minimal.

**Theorem 9:** Assume that we are using majority voting, $t = 0.5$, and that $k = 3, 5, 7, \ldots$. Suppose that:

$$(\forall i)\left[\left[\rho_i = \frac{1}{2} + \frac{1}{2k}\right] \vee \left[\rho_i = \frac{1}{2} - \frac{1}{2k}\right]\right] \tag{139}$$

This is the least stable situation; other values of $\rho_i$ are more stable. $E(\|\vec{e}_s\|)$ for $m_k$ is determined by the following formula:

$$E(\|\vec{e}_s\|) = 2nP_k - 2nP_k^2 \tag{140}$$

In this formula, $P_k$ depends on $m_k$. $P_k$ has the following values for $m_1$, $m_3$, $m_5$, and $m_\infty$:

$$P_1 = p \tag{141}$$

$$P_3 = 2p - 3p^2 + 2p^3 \tag{142}$$

$$P_5 = 3p - 9p^2 + 16p^3 - 15p^4 + 6p^5 \tag{143}$$

$$P_\infty = \begin{bmatrix} 0 & \text{if} & p = 0 \\ 0.5 & \text{if} & 0 < p < 1 \\ 1 & \text{if} & p = 1 \end{bmatrix} \tag{144}$$

*Proof:* Let $\vec{v}_1, \ldots, \vec{v}_k$ be the $k$ nearest neighbors to $\vec{v}$. Let $y_{1,1}, \ldots, y_{1,k}$ be the corresponding elements in $\vec{y}_1$. Let $y_{2,1}, \ldots, y_{2,k}$ be the corresponding elements in $\vec{y}_2$. The stability of

$m_k$ is determined by the probability that:

$$\sum_{i=1}^{k} y_{1,i} < k/2 \qquad \text{and} \qquad \sum_{i=1}^{k} y_{2,i} > k/2 \tag{145}$$

The worst case — the case with maximal expected instability $E(\|\vec{e}_s\|)$ — arises when:

$$\sum_{i=1}^{k} f(\vec{v}_i) = \frac{k+1}{2} \qquad \text{or} \qquad \sum_{i=1}^{k} f(\vec{v}_i) = \frac{k-1}{2} \tag{146}$$

Without loss of generality, we may assume that:

$$s = \frac{k+1}{2} \tag{147}$$

$$f(\vec{v}_1) = \ldots = f(\vec{v}_s) = 1 \tag{148}$$

$$f(\vec{v}_{s+1}) = \ldots = f(\vec{v}_k) = 0 \tag{149}$$

We see that:

$$y_{i,j} = \neg z_{i,j} \qquad i = 1, 2 \qquad j = 1, \ldots, s \tag{150}$$

$$y_{i,j} = z_{i,j} \qquad i = 1, 2 \qquad j = s+1, \ldots, k \tag{151}$$

Therefore:

$$P\left(\sum_{i=1}^{k} y_{1,i} > k/2\right) = P\left(\sum_{i=1}^{s} z_{1,i} \le \sum_{i=s+1}^{k} z_{1,i}\right) \tag{152}$$

We may use the term $P_k$ to indicate the probability of (152). That is, $P_k$ is the probability that the majority of $y_{1,1}, \ldots, y_{1,k}$ are 1, given (148) and (149). If the worst case holds for every row vector in $X$ — that is, (139) is true — then we have:

$$E(\|\vec{e}_s\|) = \sum_{i=1}^{n} 2P_k(1-P_k) = 2nP_k(1-P_k) \tag{153}$$

Let us consider the case $k = 3$. We have:

$$P_3 = 2p - 3p^2 + 2p^3 \tag{154}$$

For $k = 5$, we have:

$$P_5 = 3p - 9p^2 + 16p^3 - 15p^4 + 6p^5 \tag{155}$$

Again, to find the behavior $m_\infty$ of the model in the limit, we can use the Central Limit Theorem. Let $z_\alpha$ and $z_\beta$ be independent random variables with a standard normal distribution. As $k$ approaches infinity, $P_k$ approaches:

$$P(z_\alpha \le z_\beta) \tag{156}$$

Therefore:

$$P_\infty = \begin{bmatrix} 0 & \text{if} & p = 0 \\ 0.5 & \text{if} & 0 < p < 1 \\ 1 & \text{if} & p = 1 \end{bmatrix} \tag{157}$$

$$\frac{E(\|\grave{e}_s\|)}{n} = 2P_\infty(1 - P_\infty) = \begin{bmatrix} 0 & \text{if} & (p = 0) \text{ or } (p = 1) \\ 0.5 & \text{if} & 0 < p < 1 \end{bmatrix} \tag{158}$$

□ .

Theorem 9 implies that, in the worst case, $m_k$ grows increasingly unstable as $k$ increases, unless $p = 0$ or $p = 1$. Figure 3 is a plot of $E(\|\grave{e}_s\|)/n$ as a function of $p$, for the models $m_1$, $m_3$, $m_5$, and $m_\infty$.

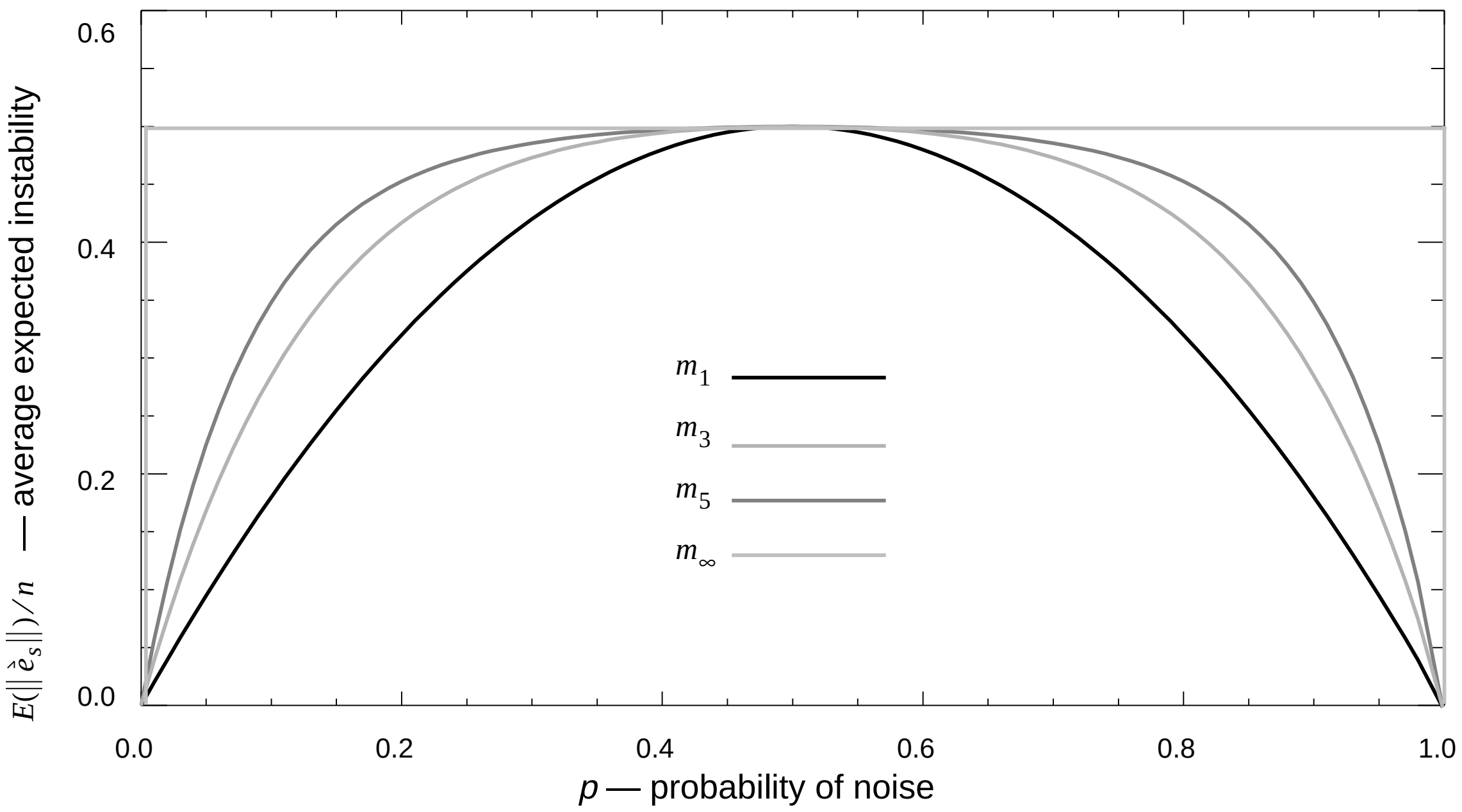

Figure 3. A plot of $E(\|\grave{e}_s\|)/n$ as a function of $p$.

## 4.3 Average Case

Sections 4.1 and 4.2 show that voting can be either very beneficial or very detrimental, depending on the relationship between the function $f$ and the voting threshold $t$. The function $f$ determines $\rho_i$, the frequency with which $f$ takes the value 1, in the neighborhood $N_k(\vec{v}_i)$ of $\vec{v}_i$. In the previous sections, we assumed majority voting, $t = 0.5$. When $\rho_i$ is near 0.5, majority voting is detrimental. When $\rho_i$ is far from 0.5, majority voting is beneficial. These results generalize to voting with other values of $t$. When $\rho_i$ is near $t$, voting is detrimental. When $\rho_i$ is far from $t$, voting is beneficial.

We would like to know how majority voting performs with an average $f$. For the average function $f$, is $\rho_i$ dangerously close to 0.5, or is it usually safely far away? This is a difficult question: What exactly is an average $f$?

The next theorem uses a simple model of an average $f$.

**Theorem 10:** Assume that we are using majority voting, $t = 0.5$. Let us suppose that we can treat $f(\vec{v}_i)$ as a random boolean variable, with probability $p'$ that $f(\vec{v}_i)$ is 1, and probability $1 - p'$ that $f(\vec{v}_i)$ is 0. In the limit, as $k$ approaches infinity, $m_k$ tends to be stable, unless either $p = 0.5$ or $p' = 0.5$.

*Proof:* We see that:

$$E\left(\sum_{i=1}^{k} f(\vec{v}_i)\right) = kp' \tag{159}$$

Recall that:

$$y_{i,j} = f(\vec{v}_j) \oplus z_{i,j} \qquad i = 1, 2 \qquad j = 1, \ldots, k \tag{160}$$

Therefore:

$$P(y_{i,j} = 0) = (1 - p')(1 - p) + p'p \tag{161}$$

$$P(y_{i,j} = 1) = p(1 - p') + p'(1 - p) \tag{162}$$

Thus:

$$E\left(\sum_{i=1}^{k} y_{1,i}\right) = k[p(1 - p') + p'(1 - p)] = \mu \tag{163}$$

$$\mathrm{var}\left(\sum_{i=1}^{k} y_{1,i}\right) = k\left[p(1-p') + p'(1-p)\right]\left[(1-p')(1-p) + p'p\right] = \sigma^2 \tag{164}$$

Consider:

$$\frac{\sum_{i=1}^{k} y_{1,i} - \mu}{\sigma} \tag{165}$$

By the Central Limit Theorem, as $k$ approaches infinity, (165) approaches a standard normal distribution. As before, we define $P_k$ as the probability:

$$P\left(\sum_{i=1}^{k} y_{1,i} > k/2\right) \tag{166}$$

Let $z_\alpha$ be a random variable with a standard normal distribution. As $k$ approaches infinity, $P_k$ approaches:

$$P\left(z_\alpha > \left(\left(k/2\right) - \mu\right)/\sigma\right) \tag{167}$$

We see that:

$$\lim_{k \to \infty} \left(\left(\left(k/2\right) - \mu\right)/\sigma\right) = \begin{bmatrix} \infty & \text{if} & [p(1-p') + p'(1-p)] < 0.5 \\ 0 & \text{if} & [p(1-p') + p'(1-p)] = 0.5 \\ -\infty & \text{if} & [p(1-p') + p'(1-p)] > 0.5 \end{bmatrix} \tag{168}$$

Therefore:

$$P_\infty = \begin{bmatrix} 0 & \text{if} & [p(1-p') + p'(1-p)] < 0.5 \\ 0.5 & \text{if} & [p(1-p') + p'(1-p)] = 0.5 \\ 1 & \text{if} & [p(1-p') + p'(1-p)] > 0.5 \end{bmatrix} \tag{169}$$

$$\frac{E(\|\grave{e}_s\|)}{n} = 2P_\infty(1-P_\infty) = \begin{bmatrix} 0 & \text{if} & [p(1-p') + p'(1-p)] \neq 0.5 \\ 0.5 & \text{if} & [p(1-p') + p'(1-p)] = 0.5 \end{bmatrix} \tag{170}$$

Thus $m_k$ tends to be stable, unless:

$$p(1-p') + p'(1-p) = 0.5 \tag{171}$$

Some algebraic manipulation converts (171) to:

$$(2p - 1)(1 - 2p') = 0 \tag{172}$$

Therefore $m_k$ tends to be stable, unless either $p = 0.5$ or $p' = 0.5$ □.

Let us now consider the general case (111), when the threshold for voting *t* is not necessarily equal to 0.5.

**Theorem 11:** Let *t* be any value, such that $0 < t < 1$. Let us suppose that we can treat $f(\vec{v}_i)$ as a random boolean variable, with probability $p'$ that $f(\vec{v}_i)$ is 1, and probability $1 - p'$ that $f(\vec{v}_i)$ is 0. Let us introduce the term $\tau$:

$$\tau = p(1 - p') + p'(1 - p) = p + p' - 2pp' \tag{173}$$

In the limit, as *k* approaches infinity, $m_k$ tends to be stable, unless $t = \tau$.

*Proof:* This is a straightforward extension of the reasoning in Theorem 10 □.

Presumably *p* (the probability that *z* is 1) and $p'$ (the probability that *f* is 1) are not under the control of the modeler. However, *t* can be adjusted by the modeler. If the modeler can estimate *p* and $p'$, then $\tau$ can be estimated. Theorem 11 implies that the modeler should make sure that *t* is relatively far from $\tau$.

The assumption that we can treat $f(\vec{v}_i)$ as a random boolean variable is somewhat unrealistic. Let us consider a more sophisticated analysis.

**Theorem 12:** Let us define $\tau_i$ as follows:

$$\tau_i = p + \rho_i - 2p\rho_i \qquad i = 1, \ldots, n \tag{174}$$

In the limit, as *k* approaches infinity, $m_k$ tends to be stable for $\vec{v}_i$, unless *t* is close to $\tau_i$.

*Proof:* This is a straightforward extension of the reasoning in Theorem 10 □.

Let us define $p'$ as the average of the $\rho_i$:

$$p' = \frac{1}{n}\sum_{i=1}^{n}\rho_i \tag{175}$$

Note that:

$$p' \neq \frac{1}{n} \sum_{i=1}^{n} f(\vec{v}_i) \qquad (176)$$

We can use (173) and (175) to calculate $\tau$. Theorem 12 implies that it is possible for $m_k$ to be stable even when $t = \tau$, if $t$ is far from each $\tau_i$.

**Theorem 13:** We can estimate $\tau_i$ from the data $(X, \vec{y}_1)$ as follows:

$$t_i = \frac{1}{k} \left( \sum_{\vec{v}_j \in N_k(\vec{v}_i)} y_{1,j} \right) \qquad i = 1, \ldots, n \qquad (177)$$

$t_i$ is an unbiased estimator for $\tau_i$.

*Proof:* We have:

$$E(t_i) = \frac{1}{k} \left( \sum_{\vec{v}_j \in N_k(\vec{v}_i)} E(y_{1,j}) \right) \qquad (178)$$

$$= \frac{1}{k} \left( \sum_{\vec{v}_j \in N_k(\vec{v}_i)} E(f(\vec{v}_j) \oplus z_{1,j}) \right) \qquad (179)$$

$$= p + \rho_i - 2p\rho_i \qquad (180)$$

$$= \tau_i \qquad (181)$$

□ .

Using (177), we can estimate $\tau_i$ for any $k$ and any $i$. $t_i$ is the frequency of the output 1, in the neighborhood $N_k(\vec{v}_i)$ of $\vec{v}_i$.

**Theorem 14:** If we know the value of $p$, then we can also estimate $\rho_i$ from the data $(X, \vec{y}_1)$:

$$r_i = \frac{t_i - p}{1 - 2p} \qquad (182)$$

$r_i$ is an unbiased estimator for $\rho_i$.

*Proof:* We have:

$$E(r_i) = \frac{E(t_i) - p}{1 - 2p} \tag{183}$$

$$= (\tau_i - p) / (1 - 2p) \tag{184}$$

$$= (p + \rho_i - 2p\rho_i - p) / (1 - 2p) \tag{185}$$

$$= \rho_i \tag{186}$$

$\square$ .

Using (177) and (182), we can estimate $\rho_i$ for any $k$ and any $i$. Once we have an estimate of $\rho_i$, we can establish whether we are closer to the best case (Section 4.1) or the worst case (Section 4.2).

We now have some understanding of the average behavior of $m_k$ in the limit. Let us consider the average behavior of $m_k$ for $k = 3$ and $t = 0.5$.

**Theorem 15:** Suppose that $n$ is even. Let $k = 3$ and $t = 0.5$. Suppose that:

$$\rho_i = 1/3 \quad \text{or} \quad \rho_i = 2/3 \quad i = 1, \ldots, n/2 \tag{187}$$

$$\rho_i = 0/3 \quad \text{or} \quad \rho_i = 3/3 \quad i = n/2 + 1, \ldots, n \tag{188}$$

Then:

$$E(\|\grave{e}_s\|) = n\,(2p - 4p^2 + 12p^3 - 26p^4 + 24p^5 - 8p^6) \tag{189}$$

*Proof:* We know from Section 4.2 that $m_3$ is unstable when:

$$\rho_i = 1/3 \quad \text{or} \quad \rho_i = 2/3 \quad i = 1, \ldots, n \tag{190}$$

We know from Section 4.1 that $m_3$ is stable when:

$$\rho_i = 0/3 \quad \text{or} \quad \rho_i = 3/3 \quad i = 1, \ldots, n \tag{191}$$

For the unstable case, $i = 1, \ldots, n/2$, we have (154):

$$P_{u,3} = 2p - 3p^2 + 2p^3 \tag{192}$$

For the stable case, $i = n/2 + 1, \ldots, n$, we have (127):

$$P_{s,3} = 3p^2 - 2p^3 \tag{193}$$

Therefore (126):

$$E(\|\grave{e}_s\|) = \sum_{i=1}^{n/2} 2P_{u,3}(1-P_{u,3}) + \sum_{i=n/2+1}^{n} 2P_{s,3}(1-P_{s,3}) \tag{194}$$

$$= \frac{n}{2} 2P_{u,3}(1-P_{u,3}) + \frac{n}{2} 2P_{s,3}(1-P_{s,3}) \tag{195}$$

$$= n\,[P_{u,3} - P_{u,3}^2 + P_{s,3} - P_{s,3}^2] \tag{196}$$

$$= n\,(2p - 4p^2 + 12p^3 - 26p^4 + 24p^5 - 8p^6) \tag{197}$$

□ .

Figure 4 is a plot of $E(\|\grave{e}_s\|)/n$ for $m_3$, using the average case analysis of Theorem 15. For comparison, $E(\|\grave{e}_s\|)/n$ is also plotted for $m_1$. $m_3$ is slightly more stable than $m_1$, except at 0, 0.5, and 1, where they are equal.

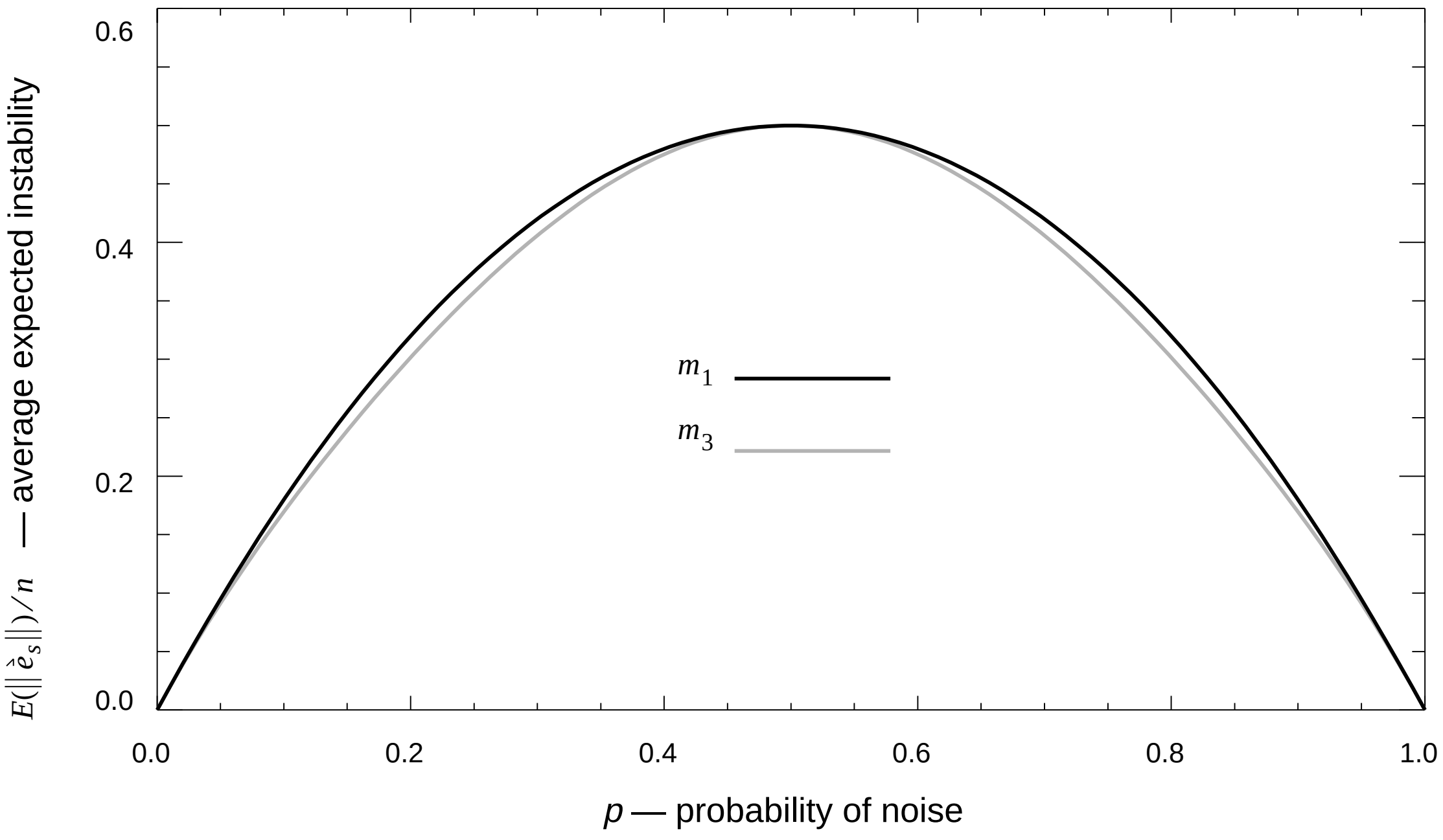

Figure 4. A plot of $E(\|\grave{e}_s\|)/n$ as a function of $p$.

## 5 Application

Let us assume that *n*, *p*, *X*, and $\grave{y}$ are fixed, while *k* and *t* are under our control. Our goal is to minimize the expected cross-validation error. The data *X* and $\grave{y}$ are the training set, and the testing set is not yet available to us. How do we determine the best values for *k* and *t*?

Using Theorems 2 or 3, we can estimate $E(\|\vec{e}_c\|)$ if we can estimate $E(\|\vec{e}_t\|)$ and $E(\|\vec{e}_s\|)$.

Obtaining an estimate of $E(\|\vec{e}_t\|)$ is relatively simple. We can use the actual error on the training set $\|\vec{e}_t\|$ as an estimator for the expected error on the training set $E(\|\vec{e}_t\|)$. By the Law of Large Numbers (Fraser, 1976), as $n$ increases, $\|\vec{e}_t\| / n$ becomes increasingly close to $E(\|\vec{e}_t\|) / n$.

Obtaining an estimate of $E(\|\vec{e}_s\|)$ is more difficult. Suppose $k = 3$, $t = 0.5$, and the value of $p$ is known by the modeler. We can use Theorem 14 to estimate $\rho_i$ for $i = 1, \ldots, n$. We can then use the method of Theorem 15 to obtain an estimate of $E(\|\vec{e}_s\|)$.

To estimate $\rho_i$, we need to know *p*, the level of noise in the black box. We can estimate *p* empirically from the data (by assuming that *f* is usually constant within a small neighborhood, for example), or we can theoretically analyze the internal mechanism of the black box. A mixed approach, combining theoretical and empirical analysis, may be fruitful.

This approach to estimating $E(\|\vec{e}_s\|)$ is feasible for $k = 3$ or $k = 5$, but the algebra becomes exponentially messy for larger values of *k*. For large values of *k*, we can use Theorem 13 to estimate $\tau_i$. Theorem 12 suggests that $E(\|\vec{e}_s\|)$ will tend toward 0, as long as $\tau_i$ is far enough from *t*, for all *i*.

Once we have an estimate of $E(\|\vec{e}_t\|)$ and $E(\|\vec{e}_s\|)$, we can use Theorem 2 to set an upper bound on $E(\|\vec{e}_c\|)$. Alternatively, we can use Theorem 3 to estimate $E(\|\vec{e}_c\|)$. (It is conjectured that $m_k$ satisfies the assumption of Theorem 3, that $\vec{e}_t$ and $\vec{e}_s$ are statistically independent, but there is not yet a proof for this.) We can estimate $E(\|\vec{e}_c\|)$ for varying values of *k* and *t*. We should then choose the values of *k* and *t* that minimize our estimated expected cross-validation error.

The average-case analysis of the previous section suggests that increasing *k* will tend to increase stability. Increasing *k* will also tend to increase training set error. If we plot cross-validation error as a function of *k*, then we expect to see a curve with a single minimum at the optimal value of *k*, somewhere between the extremes $k = 1$ and $k = n$. Empirical tests of nearest neighbor algorithms report exactly this result (Dasarathy, 1991). I have not included any empirical tests with real-world data in this paper, because it would merely duplicate work that has already been done. That is, the average-case analysis presented here is consistent with typical empirical experience.

The worst-case analysis, on the other hand, yields a surprising result, that increasing $k$ can actually *decrease* stability. In the worst case, if we plot cross-validation error as a function of $k$, then we may see a curve with a minimum at $k = 1$, that steadily rises as $k$ increases. It is sometimes reported that cross-validation error is minimal at $k = 1$, but it is usually assumed that this is due to rapid increase in training set error with increasing $k$, rather than decrease in stability with increasing $k$. It seems likely that the worst case, where stability decreases as $k$ increases, is very rare in real-world data. Thus it would be difficult to find real-world data that illustrates the worst-case analysis. Simulated data could be used, but there would be no point in such an exercise, since the results would be determined by the assumptions embedded in the simulation. Therefore this paper does not include any empirical results.

# 6 Related Work

This work is most closely related to Turney (1993), which presents a general theory of cross-validation error for algorithms that predict real-valued attributes. The theory is then applied to linear regression and instance-based learning. The theory of cross-validation error for algorithms that predict boolean-valued attributes, presented in Section 2, is similar to the theory for real-valued attributes (Turney, 1993), but there are some differences. Real-valued noise is quite different from boolean-valued noise. Boolean noise $z$ is a random variable in $\{0,1\}$, such that the probability that $z$ is 1 is $p$. Real noise $z$ is a random variable in $\Re$, such that $z$ has mean 0 and variance $\sigma^2$. Equivalently, the noise is $\sigma z$, where $z$ has mean 0 and variance 1. The role of $p$ in boolean noise is similar to the role of $\sigma$ in real noise, except that $p$ must be in the range $[0,1]$, but the range of $\sigma$ is $\Re^+$. This makes a substantial difference in the details of the two cases, real-valued and boolean-valued.

Turney (1993) also examined the cross-validation error of instance-based learning with $k$ nearest neighbors. When instance-based learning is used to predict real-values, the $k$ nearest neighbors are averaged together (Kibler *et al.*, 1989). Averaging the $k$ nearest neighbors for real-valued attributes is analogous to voting for boolean-valued attributes. There is a surprising difference, however. Turney (1993) proves that averaging, in the best case, improves stability and, in the worst case, does not affect stability (Theorem 10 in (Turney, 1993)). Voting, in the worst case, can be *destabilizing* (Section 4.2). It is surpris-ing to discover this break in the analogy between the real-valued case and the boolean-valued case.

Turney (1993) mentions that the theory of cross-validation error is similar to the work in Akaike Information Criterion (AIC) statistics (Sakamoto *et al.*, 1986). The similarity stems from the idea of finding the best model $\vec{m}(X|X,\vec{y}_1)$ for one set of outputs $\vec{y}_1$, then evaluating the model's performance with a second set of outputs $\vec{y}_2$. This is the definition of the expected cross-validation error. Akaike uses the same approach to define *mean expected log likelihood* (Sakamoto *et al.*, 1986). For a more thorough comparison of the approach described here with AIC, see Turney (1993).

There is an interesting connection between the work here and previous work in nearest neighbor pattern classification. Cover and Hart (1967) assume that *n* is large and the probability distribution for $P(f(\vec{v}) \oplus z = 1)$ is a continuous function of $\vec{v}$. Using Lemma 1, they prove the following Theorem:

$$R^* \le R \le R^* \left(2 - MR^* / (M-1)\right) \tag{198}$$

Here, *R* is the probability of error of $m_1$, $R^*$ is the Bayes probability of error, and *M* is the number of categories. The Bayes probability of error is the minimum probability of error over all decision rules taking underlying probability structure into account (Cover & Hart, 1967). This paper assumes boolean-valued attributes, so $M = 2$:

$$R^* \le R \le 2R - 2(R^*)^2 \tag{199}$$

Compare this with Theorem 6:

$$p_\beta = 2p_\alpha - 2p_\alpha^2 \tag{200}$$

There are clearly some similarities between (199) and (200). However, there are also some differences. (200) is an exact equality, while (199) is an inequality. The reason for this difference is that Cover and Hart's (1967) noise, $z(\vec{v})$, is a function of $\vec{v}$, while the noise *z* discussed here is not a function of $\vec{v}$. To remove the dependence on $\vec{v}$, Cover and Hart (1967) define *R* and $R^*$ as the expected probabilities, where the expectation is taken over $\vec{v}$. This makes their proof more complex than mine.

Cover and Hart's (1967) Lemma is similar to Theorem 1 here. Their Theorem is similar to Theorem 6 here. They do not have any results comparable to Theorems 2 and 3. That is, they do not separate cross-validation error into two components, accuracy and stability. It is this separation which enables me to achieve the results of Section 4. Section 4 does not correspond to any previous work with nearest neighbor pattern classification. Section 4 is evidence in favor of the two-component view of cross-validation error.

We know from Section 3 that $m_1$ has maximal accuracy, since $E(\|\vec{e}_t\|) = 0$. In general, $m_k$ will be less accurate than $m_1$, but $m_k$ will also be more stable than $m_1$. As $k$ increases, we expect that accuracy decreases (error on the training set increases) and stability increases. At some point, as $k$ increases, further increase in stability will not sufficiently compensate for the decrease in accuracy. Section 5 shows how we can find the value of $k$ that gives the best balance between accuracy and stability. It is certainly possible, for a certain set of data, that the best balance is $k = 1$. This is not a new insight. Cover and Hart (1967) prove that there are situations in which $m_1$ has a lower probability of error than $m_k$, where $k > 1$. What is new is the observation that there are situations in which, as $k$ increases, stability may *decrease* (see Section 4.2). Thus both accuracy and stability decrease. Section 5 shows how we may detect such situations, using the estimators $r_i$ for $\rho_i$ and $t_i$ for $\tau_i$.

Aha *et al.* (1991) also examine $k$ nearest neighbor algorithms, but their analysis is a worst-case analysis, assuming certain probability distributions. They also assume numeric-valued attributes.

Langley (1993) also presents an average-case analysis of a nearest neighbor algorithm. Langley's (1993) analysis assumes a conjunctive target concept, noise-free boolean attributes, and a uniform distribution over the instance space. These assumptions do not overlap with the assumptions made here, which makes the two analyses highly complementary. A worthwhile project for the future would be to integrate the two analyses in a single framework.

# 7 Future Work

The weak point of this theoretical model of cross-validation error is the assumption that the inputs $X$ are the same in the training set $(X, \vec{y}_1)$ and the testing set $(X, \vec{y}_2)$. The reason for making this assumption is that it simplifies the mathematics. If the inputs in the training set were different from the inputs in the testing set, then we would need to make some assumptions about $f$ before we could prove any interesting results. When the inputs are the same in the training and testing sets, we can prove some interesting results about cross-validation error, without making any assumptions about the deterministic component *f* of the black box.

The main implication of the assumption, that *X* is the same in the training set and the testing set, is that we may be underestimating the cross-validation error. We can expect a model to have more difficulty with prediction when the inputs in the testing set are different from the inputs in the training set. Theorem 1 shows that the assumption is reasonable for large *n*, but we would like to know what happens when *n* is small. When a learning algorithm makes predictions for inputs in the testing set that do not appear in the training set, the algorithm is performing interpolation or extrapolation. Future work should extend the theory of cross-validation error to cover interpolation and extrapolation.

Another area for future work is proving that instance-based learning satisfies the assumptions of Theorem 3. It is conjectured that Theorem 3 is true for instance-based learning, or that it is true under certain weak conditions. However, there is not yet a proof for this.

Finally, it would be interesting to apply the general theory of cross-validation error to some other learning algorithms, such as ID3. Even when analytical results are not feasible, it is possible to use Monte Carlo techniques to evaluate the stability of learning algorithms. A Monte Carlo approach may also be required to evaluate the stability of some of the more elaborate instance-based learning or memory-based reasoning algorithms. The complexity of some of these algorithms makes a formal analysis difficult.

# 8 Conclusion

This paper introduces a general theory of cross-validation error for algorithms that predict boolean-valued attributes. It shows that cross-validation error has two components, accuracy and stability. The optimal cross-validation error is derived and it is shown that algorithms that maximize accuracy will give sub-optimal cross-validation error.

The theory was then applied to an analysis of voting in instance-based learning. It was proven that single nearest neighbor algorithms are unstable, because they maximize accuracy. The motivation for voting is to increase stability. In the best case, voting does indeed increase stability. However, in the worst case, voting can actually decrease stability. Techniques were provided for estimating the expected cross-validation error. These techniques can be used to find the best values for *k* (the size of the neighborhood) and *t* (the threshold for voting).

# Acknowledgments

I would like to thank Dr. David Aha for extensive, helpful comments on this paper. I would also like to thank Professor Malcolm Forster for pointing out the relation between my work and Akaike Information Criterion statistics.

# Notes

1. The work described here does not rely on any of the results in Turney (1993). However, to avoid duplication, some relevant issues that are discussed in Turney (1993) are not discussed here (and vice versa). Therefore, although the work here stands on its own, the interested reader may wish to also read Turney (1993).

2. Weiss and Kulikowski (1991) reserve the term "cross-validation" for $N$-fold cross-validation. In $N$-fold cross-validation, the data set is split into $N$ subsets of roughly equal size. The classification algorithm is then tested $N$ times, each time training with $N - 1$ of the subsets and testing with the remaining subset. Therefore $N$-fold cross-validation is essentially $N$ applications of what is called "cross-validation" here. The essential concepts remain the same whether the process is done once or $N$ times.

3. It is a common practice in statistics to prove theorems that hold *asymptotically*, as the size of the data set increases to infinity. For example, Cover and Hart (1967) prove asymptotic theorems. Informally, we may say that asymptotic theorems assume an infinite-sized data set. The first assumption here, that the inputs are the same in the testing and training sets, is closely related to the assumption that the data sets are infinitely large. This is the point of Theorem 1 and Lemma 1. However, assumption 1 may be satisfied by small, finite sets of data, when the data are generated by a planned, controlled series of experiments. Thus assumption 1 is weaker than the assumption of infinite data sets.

# Appendix: List of Symbols

Note: Symbols that only occur in one theorem are omitted from this list.

| | |
|---|---|
| $\ldots \oplus \ldots$ | the exclusive-or operator |
| $\lVert \ldots \rVert$ | the length of a vector |
| $\mu$ | the mean of a probability distribution |
| $\rho_i$ | the frequency with which $f$ takes the value 1 in $N_k(\vec{v}_i)$ |
| $\sigma$ | the standard deviation of a probability distribution |
| $\sigma^2$ | the variance of a probability distribution |
| $\tau$ | the value of the threshold $t$ at which voting becomes unstable |
| $\tau_i$ | $\tau$ for a particular neighborhood $N_k(\vec{v}_i)$ |
| $\vec{e}_c$ | the cross-validation error vector |
| $\vec{e}_s$ | the instability vector |
| $\vec{e}_t$ | the training set error vector |
| $E(\ldots)$ | the expectation operator |
| $f(\vec{v})$ | a scalar function of the input vector $\vec{v}$; the target concept |
| $\vec{f}(X)$ | a vector function of the input matrix $X$ |
| $k$ | the size of the neighborhood |
| $m(\vec{v} \mid X, \vec{y})$ | the model's prediction for $f(\vec{v})$, given data $X$ and $\vec{y}$; a scalar |
| $\vec{m}(X \mid X, \vec{y})$ | the model's prediction for $\vec{f}(X)$, given the data $X$ and $\vec{y}$; a vector |
| $m_\alpha$ | the optimal model, assuming that $f$ and $p$ are known to the modeler |
| $m_\beta$ | the model $\vec{m}(X \mid X, \vec{y}_i) = \vec{y}_i$ |
| $n$ | the number of samples; the number of observations |
| $N_k(\vec{v}_i)$ | the set of the $k$ nearest neighbors $\vec{v}_1, \ldots, \vec{v}_k$ to $\vec{v}_i$ |
| $p$ | the probability that a random boolean variable $z$ has the value 1 |
| $p'$ | the probability that $f(\vec{v})$ is 1, when $f(\vec{v})$ is treated as a random variable |
| $p_\alpha$ | the probability of error $E(\|\vec{e}_c\|)/n$ for $m_\alpha$ |
| $p_\beta$ | the probability of error $E(\|\vec{e}_c\|)/n$ for $m_\beta$ |
| $P_k$ | the probability that the majority of $y_{1,1}, \ldots, y_{1,k}$ are 1 |
| $r$ | the number of inputs; the number of features |
| $r_i$ | an unbiased estimator for $\rho_i$ |
| $R$ | $p_\beta$ in the terminology of Cover and Hart (1967) |
| $R^*$ | $p_\alpha$ in the terminology of Cover and Hart (1967) |
| $\mathrm{sim}(\ldots,\ldots)$ | the similarity measure; a scalar measure of similarity between vectors |
| $t$ | a threshold for voting, where $0 < t < 1$; with majority voting, $t = 0.5$ |
| $t_i$ | an unbiased estimator for $\tau_i$ |
| $\vec{v} = \begin{bmatrix} x_1 & \ldots & x_r \end{bmatrix}$ | a vector of inputs; the attributes of an observation; features |
| $\vec{v}_i$ | the $i$-th row in $X$; the $i$-th input vector; the $i$-th observation |
| $X$ | a matrix of inputs; rows are observations and columns are features |
| $y$ | the output; the class variable |
| $z$ | a random boolean variable; noise in the class variable $y$ |
| $\vec{z}$ | a vector of random boolean variables |

# Figures

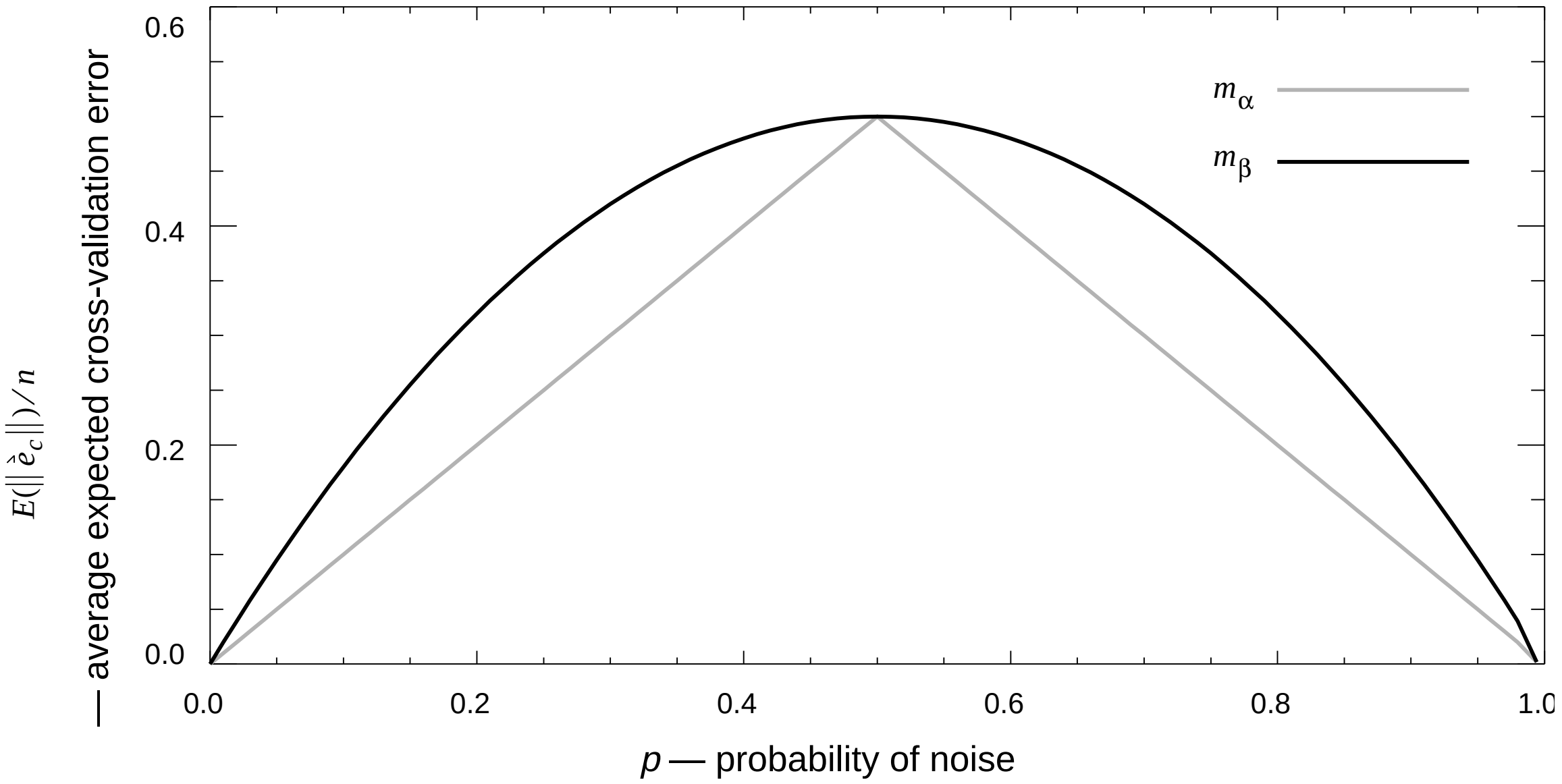

Figure 1. Plot of $E(\|\grave{e}_c\|)/n$ as a function of $p$.

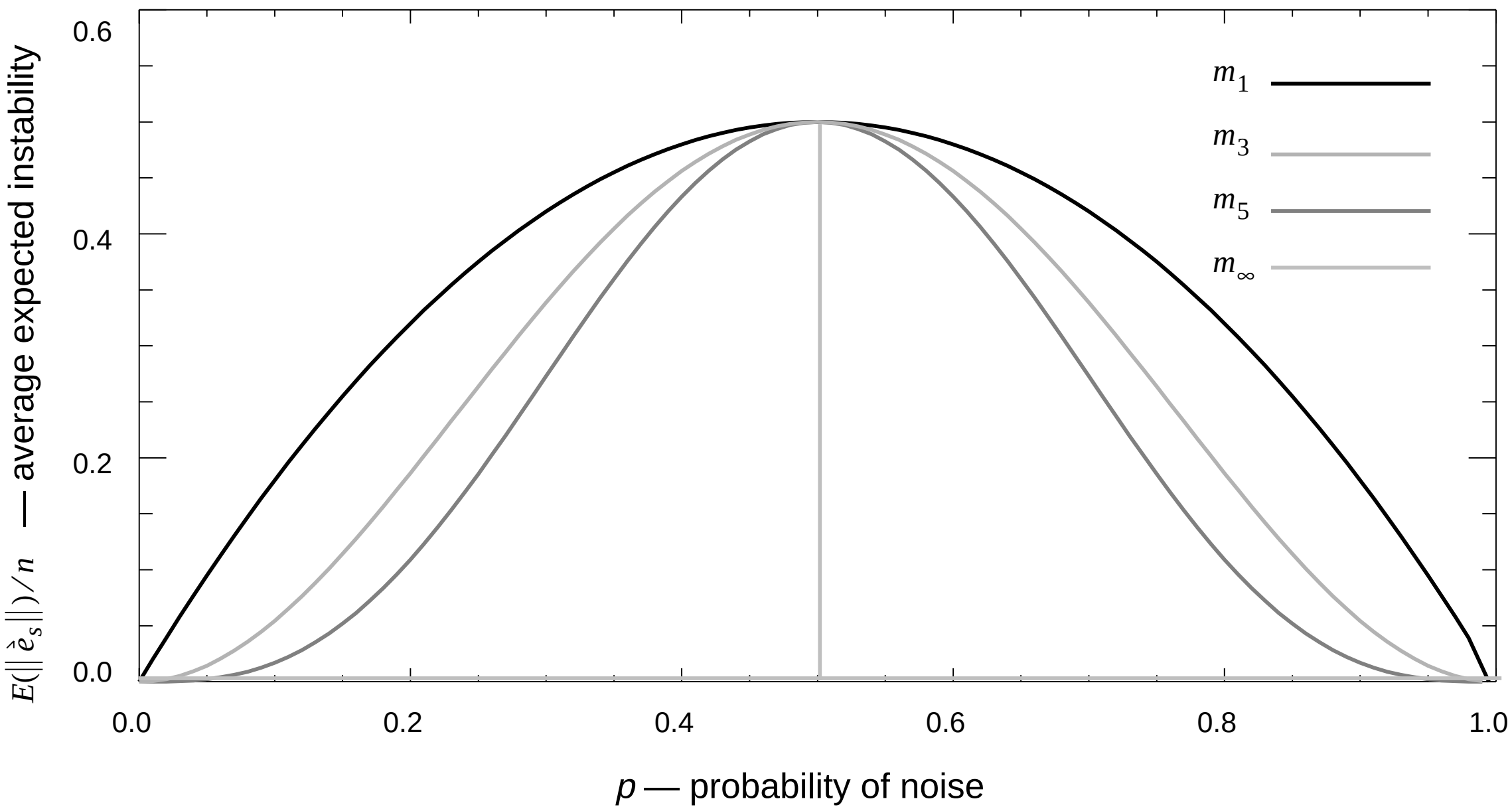

Figure 2. A plot of $E(\|\grave{e}_s\|)/n$ as a function of $p$.

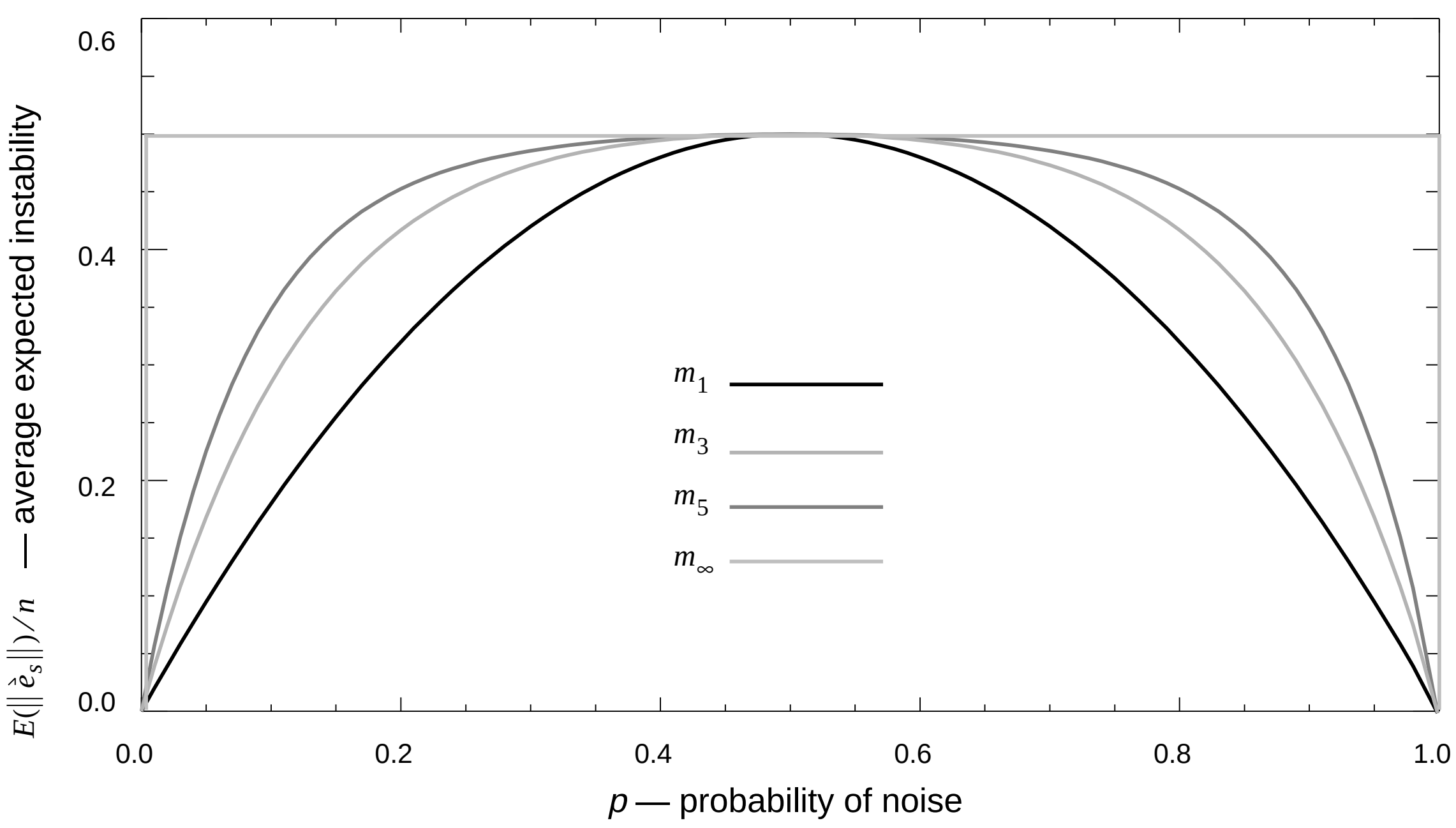

Figure 3. A plot of $E(\|\grave{e}_s\|)/n$ as a function of $p$.

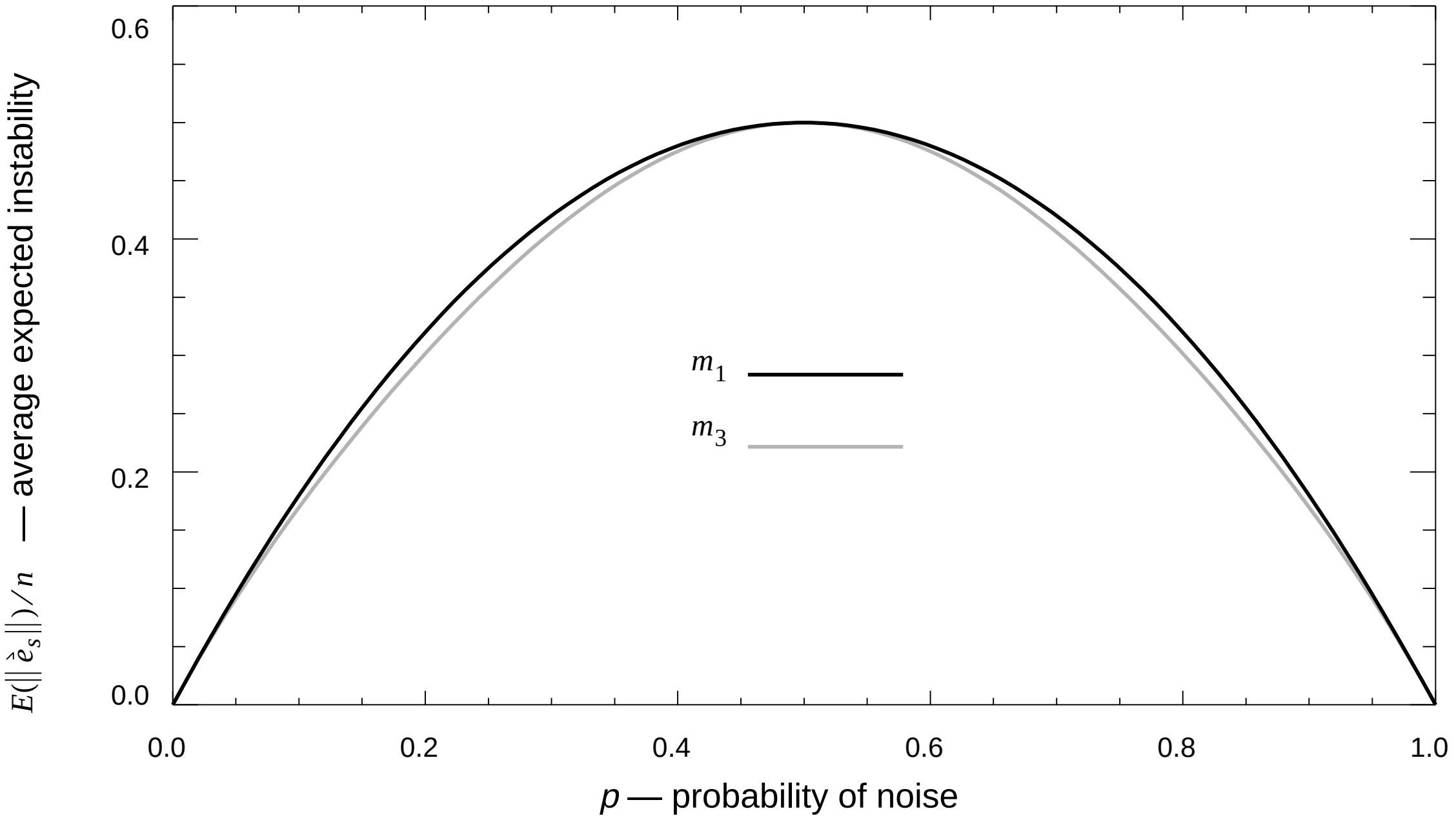

Figure 4. A plot of $E(\| \vec{e}_s \|)/n$ as a function of $p$.